\documentclass[10pt,twocolumn,letterpaper]{article}

\usepackage{iccv}
\usepackage{times}
\usepackage{epsfig}
\usepackage{graphicx}
\usepackage{amsmath}
\usepackage{amssymb,amsbsy}
\usepackage{array}
\usepackage{subfig}
\usepackage{mathtools}
\usepackage{enumitem}
\usepackage{url}

\newcommand*\Bell{\ensuremath{\boldsymbol\ell}}

\setlist[itemize]{leftmargin=*}

\usepackage[pagebackref=true,breaklinks=true,letterpaper=true,colorlinks,bookmarks=false]{hyperref}

\iccvfinalcopy 

\def\iccvPaperID{1444} 

\begin{document}

\title{Guided Perturbations: Self Corrective Behavior in Convolutional Neural Networks}

\author{Swami Sankaranarayanan \thanks{Work performed while interning at GE Global Research}\\
University of Maryland\\
College Park, MD\\
{\tt\small swamiviv@umiacs.umd.edu}
\and
Arpit Jain\\
GE Global Research Center\\
Niskayuna, NY\\
{\tt\small Arpit.Jain@ge.com}
\and
Ser Nam Lim\\
GE Global Research Center\\
Niskayuna, NY\\
{\tt\small limser@ge.com}
}

\maketitle

\begin{abstract}

Convolutional Neural Networks have been a subject of great importance
over the past decade and great strides have been made in their utility
for producing state of the art performance in many computer vision
problems. However, the behavior of deep networks is yet to be fully
understood and is still an active area of research.  In this work, we present an intriguing behavior: pre-trained CNNs can be made to improve their predictions by structurally perturbing the input. We observe
that these perturbations - referred as Guided Perturbations - enable a trained network to improve its prediction
performance without any learning or change in network weights. We perform various ablative experiments to understand how these perturbations affect the local context and feature representations. Furthermore, we demonstrate that this idea can improve performance of several existing approaches on semantic segmentation and scene labeling tasks on the PASCAL VOC dataset and supervised classification
tasks on MNIST and CIFAR10 datasets.

\end{abstract}

\section{Introduction}\label{sec:intro}
\vspace{-2mm}
Convolutional Neural Networks (CNNs) have achieved state of the art
results on several computer vision benchmarks such as ILSVRC
\cite{ILSVRC15} and PASCAL VOC \cite{pascalvocdb} over the past few
years. Despite their overwhelming success, recent results have
highlighted that they can be sensitive to small adversarial noise in
the input \cite{adv2014} or can be easily fooled using structured noise
patterns \cite{nguyen2015deep}. To understand how a CNN can learn
complex and meaningful representations but at the same time be easily
fooled by simple and imperceptible perturbations still remains an open
research problem. The work of Goodfellow \emph{et. al.} \cite{adv2014}
and Szegedy \emph{et al.} \cite{intriguing} among others, bring out
the intriguing properties of neural networks by introducing
perturbations in either the hidden layer weights or the input image.
While these approaches have focused on understanding the effect of
adversarial noise in deep networks, in this work we present an
interesting observation: input perturbations can enable a CNN to
correct its mistakes. We find that these perturbations exploit the local neighborhood information from network's prediction which in turn results in contextually smooth predictions.

\begin{figure}[!t]
\centering
\includegraphics[width=0.45\textwidth, height=0.45\textwidth]{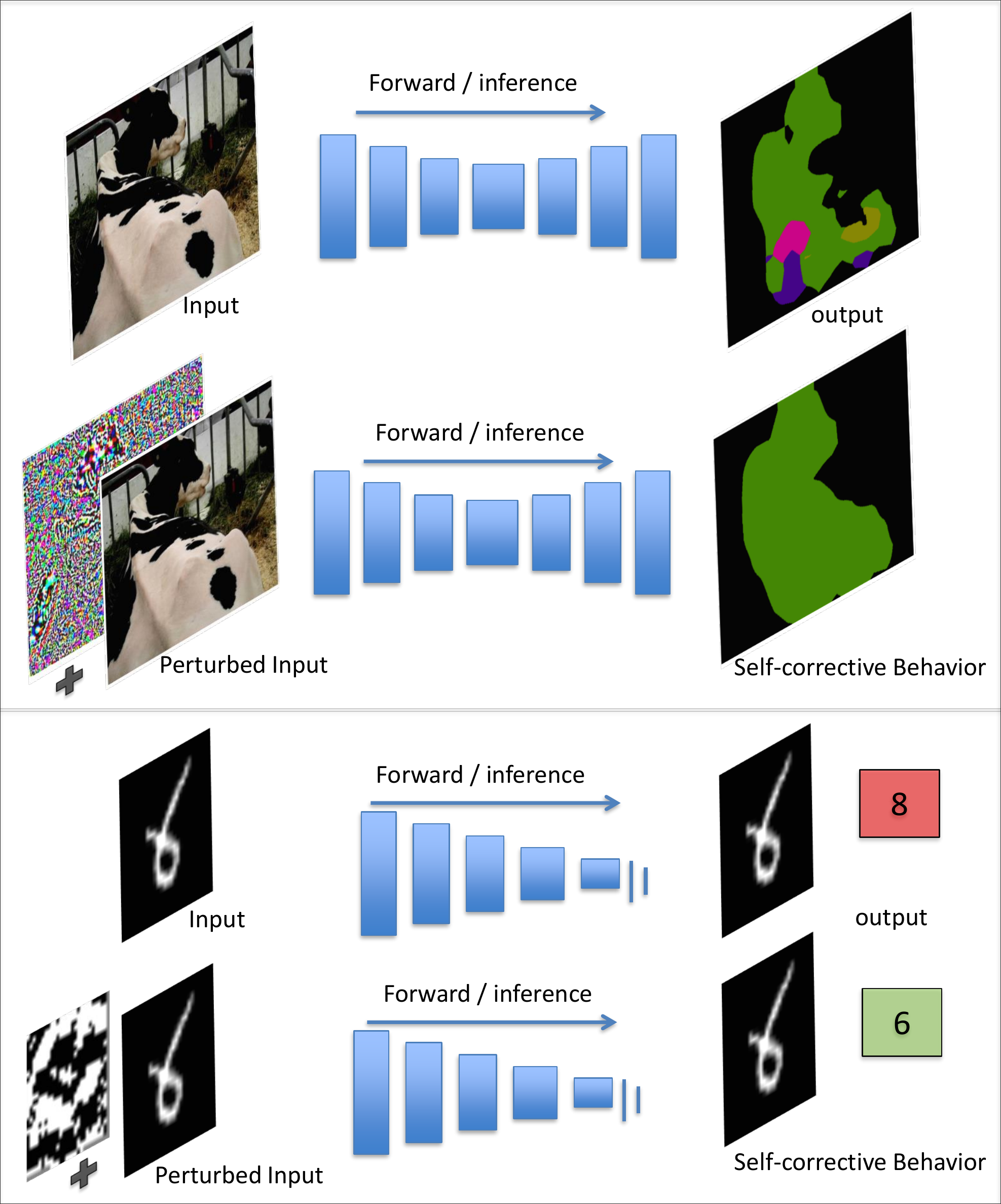} 
\caption{Self-corrective behavior due to Guided Perturbations for segmentation and classification tasks.}
\label{fig:cow}
\vspace{-6mm}
\end{figure}

In almost all the CNN based approaches, the output is obtained using a single forward pass during the prediction time. In the proposed approach, we use the prediction made by the network during the forward pass to generate perturbations at the input. Specifically, we backpropagate the gradient of the prediction error all the way to the input. We would like to emphasize that the error gradients are generated purely based on the network's prediction without any knowledge of ground truth. We \textit{perturb} the input image by adding to it a scaled version of the gradient signal. This is fed back to the network again for prediction. Figure \ref{fig:cow} shows an example of the self-corrective behavior of the generated perturbations for segmentation and classification tasks. This example shows that these perturbations of the input image could be viewed as a form of structured distortion that is added to the input such that the context gets amplified in each pixel's neighborhood which enables the network to correct its own mistakes. The proposed approach is simple and easy to implement and does not require retraining or modification in network's architecture. 

Existing approaches to improve performance on segmentation and classification tasks have been geared towards novelties in network architecture or using large amount of training data or both. While these are valid ways to improve the network’s performance, the proposed approach highlights an inherent behavior of CNNs that can be used to improve their prediction without requiring additional learning or training data. We would like to note here that while the behavior of Guided Perturbations (GP) is similar to Conditional Random Fields based approaches, the difference in our case is that there is no explicit modeling of context or neighborhood interactions. Since our approach is network independent, this doesn't preclude networks which model context explicitly and we show improvements in such networks too.



To the best of our knowledge, this is the first approach to show existence of a self-corrective behavior in CNNs and use of such behavior for improvement in performance on segmentation and classification tasks. To summarize, the major contributions of this paper are:

\begin{itemize}
\vspace{-2mm}
\item We present a novel and intriguing observation: there exist structured perturbations which when used to perturb the input leads to a corrective behavior in CNNs. 
\vspace{-2mm}
\item We propose a generalized framework to improve the performance of any pretrained CNN model that is architecture independent and requires no learning assuming the network is trained end-to-end.  

\end{itemize}

\section{Background}\label{sec:bg}
In recent years, there have been several approaches that attempt to analyze the behavior of CNNs for classification problems. Mahendran~\etal \cite{inversion2015} proposed an approach to invert
the function learned by the CNN in order to generate as faithful a
reconstruction of the input as possible. This is performed by
minimizing a regularized energy function that approximates the
representation function that is learned by the deep network. Another
interesting work in this direction is the Fooling Images work of Nguyen~\etal
\cite{nguyen2015deep} that is further extended by Yosinski~\etal
\cite{deepvis}. The main objective in both the approaches is to
synthesize images to confuse CNN by maximizing the activation of individual neurons from different layers of a deep network. This leads to interesting
results such as images that look like random noise but which the CNN
classifies into different classes with high confidence. The approaches
that are closer in spirit to our proposed approach are the ones
that predate these recent ones: Szegedy~\etal \cite{intriguing} and
Goodfellow~\etal \cite{adv2014}. Their study shows
that there exist a lot of adversarial examples which are the result of
minor pertubations of the input that causes the CNN to misclassify
input images on classification tasks; these examples can be generated
by adding a fraction of the gradient that is generated by wiggling the
classifier output in the direction of the target class. 

One of the applications that this paper focuses on is semantic segmentation. A lot of research have gone into understanding the expressive ability of CNNs for such problems. Recent methods for image segmentation such as Fully Convolutional Networks (FCN) by Long~\etal \cite{fcncvpr} have provided an easy framework that casts the image segmentation problem as a pixelwise label classification
problem 
The major difference in their work was the image level output
generation and backpropogation which was made possible by the work of
Zeiler~\etal \cite{deconv}. This image level back propogation provides
a simple way to learn a discriminative representation of classes at
the pixel level. Several recent approaches such as \textit{CRFasRNN}
by Zheng~\etal \cite{crfasrnn}, DeepLab by
Chen~\etal\cite{chen2016deeplab} and GCRF by Vemulapalli~\etal
\cite{gcrf} have improved the FCN framework by explicitly modeling context. CRFasRNN casts the CRF iterations, which has been
traditionally used as a post processing function in image segmentation
problem to ensure label compatibility, as a Recurrent Neural Network.
They formulate the steps required to perform a mean field iteration in
a CRF including message passing and learning a label compatibility
transform as a layer in a CNN, which is unrolled in time over $T$
iterations. The unary potentials are computed using the FCN-8s network
which is then refined using the RNN strucure. By casting this as a CNN
layer they perform end-to-end training. 
More recently, Yu~\etal \cite{mscontext} propose to train a
multiscale context aggregation module on top of a modified FCN-8s
network. This context module improves the performance of the base
network on its own or in combination with CRF based approaches.

In this paper, we describe an interesting property of deep networks about how
it can change its predictions for the better using minor perturbations of the input. Furthermore, we provide useful applications of our approach by showing how it can be used to improve prediction performance on challenging computer vision tasks.
\section{Our Approach}\label{sec:method}

\begin{figure}[!b]
\centering
\includegraphics[width=0.5\textwidth]{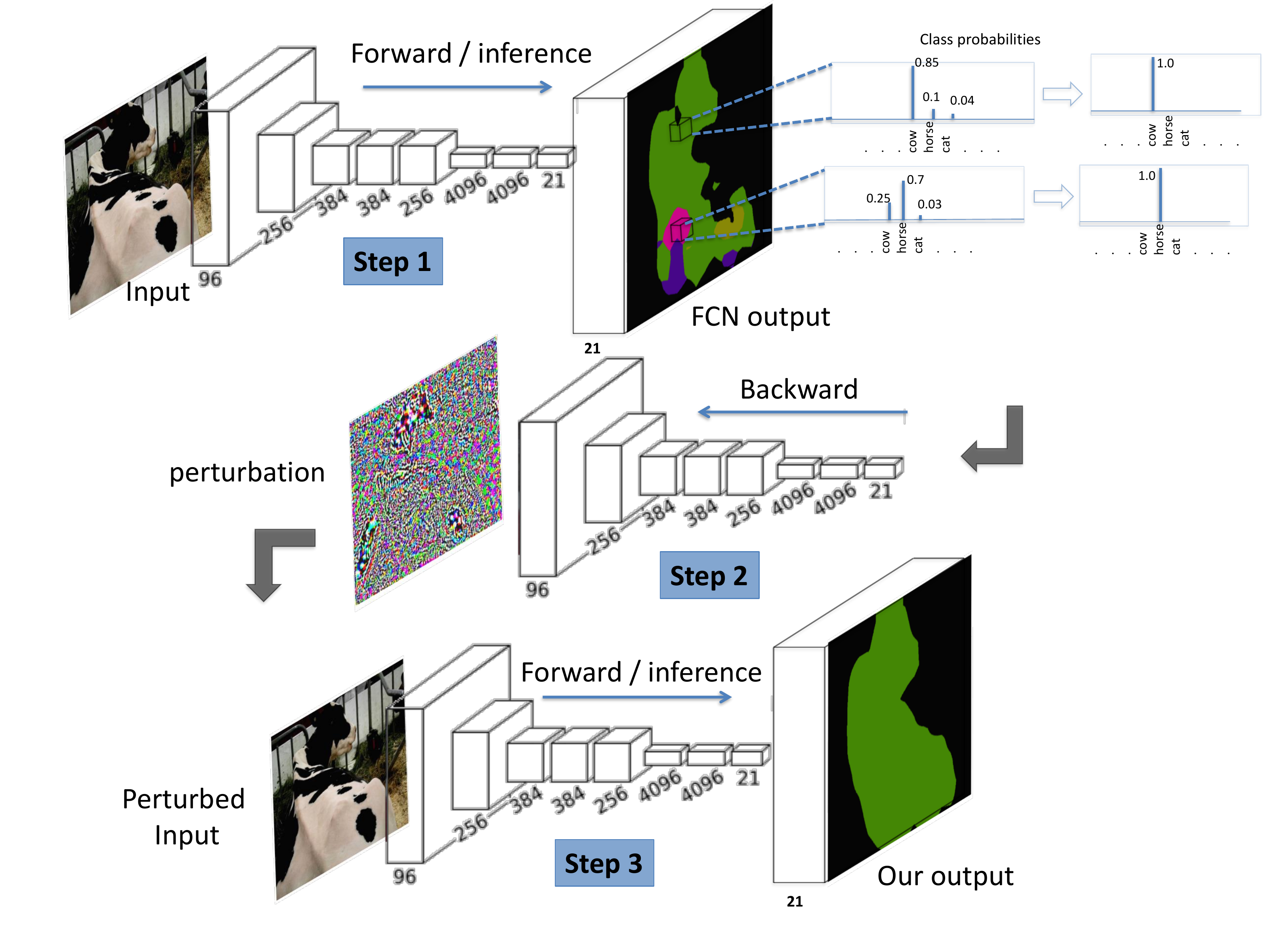} 
\caption{Processing pipeline for the proposed approach for semantic segmentation}
\label{fig:approach}
\vspace{-3mm}
\end{figure} 

In this section, we describe our approach to generate guided perturbations by using the gradient information obtained from the network's output. We perform experiments to study different aspects of these perturbations and how they affect the network representations. Since our approach to generate guided perturbation is different for segmentation and classification tasks, we discuss them separately. 

\subsection{Semantic Segmentation}\label{subsec:segment}
Figure \ref{fig:approach} illustrates our approach for semantic segmentation task. Given an input image we perform a forward pass to compute the output - which is usually the output of a softmax function that gives a class probability vector for each pixel. The prediction output is then binarized by setting the probability of the most confident class to one and the others to zero. This is done for each pixel and the error gradient is computed at the softmax layer by setting this modified output as ground truth.
Let $\mathbb{X} \in \mathcal{R}^{M \times N \times C_{in}}$ represent the input image to the deep network, $\mathbb{Y} \in \mathcal{R}^{M \times N \times C_{out}}$ represent ground truth labeling, where $C_{in} $ is the number of input channels, $C_{out}$ is the number of classes and $M  \times N$ is the dimensionality of the input image. Let $\theta$ represent the parameters of the network and  $\mathcal{J(\theta,\mathbb{X},\mathbb{Y})}$ represent the loss function that is optimized during training. During prediction time, let $\mathbb{Y}_{pred}$ be the predicted labeling. In order to generate an error gradient for backpropagation, we create a pseudo ground truth labeling $\mathbb{Y}_{pseudo}$ by modifying $\mathbb{Y}_{pred}$ as follows: We initialize $\mathbb{Y}_{pseudo}$ with $\mathbb{Y}_{pred}$. Let the $k^{th}$ component of $\mathbb{Y}_{pseudo}$ be represented as $\mathbf{y_k}=[y_{k_1},..,y_{k_{C_{out}}}]$, which is a $C_{out}$-dimensional score vector. We modify $\mathbf{y_k}$ to be a 1-hot encoded vector with the maximally confident class set to 1 and others to zero. Then, the error gradient signal is computed based on the loss function $\mathcal{J}(\theta,\mathbb{X},\mathbb{Y}_{pseudo})$ and backpropagated through the network up to the input. Let the backpropagated error gradient signal at the input be represented as: $\nabla_X \mathcal{J}(\theta,\mathbb{X},\mathbb{Y}_{pseudo})$. The perturbed input in then generated as follows:
\begin{equation}
\mathbb{X}_{per}=\mathbb{X}+\epsilon sign(\nabla_X \mathcal{J}(\theta,\mathbb{X},\mathbb{Y}_{pseudo})), \qquad \epsilon > 0
\end{equation}



 
 where $\epsilon$ is a non negative scaling factor that is model dependent and $sign(.)$ represents the signum function taken elementwise. $\mathbb{X}_{per}$ is then fed into the network for a forward pass to generate the final output. 

It can be argued that the above method of generating gradients using the network's prediction can lead to inaccurate gradient information propagated through the network especially in cases where the network's output contains many misclassified pixels. The key insight we provide in this work is that despite misclassifications by the deep network, the gradients at the input obtained from the network's prediction, in general, improve the final output of the deep network.


\begin{figure}[!htb]
\vspace{-1mm}
\centering
\includegraphics[width=0.5\textwidth,height=0.65\textwidth]{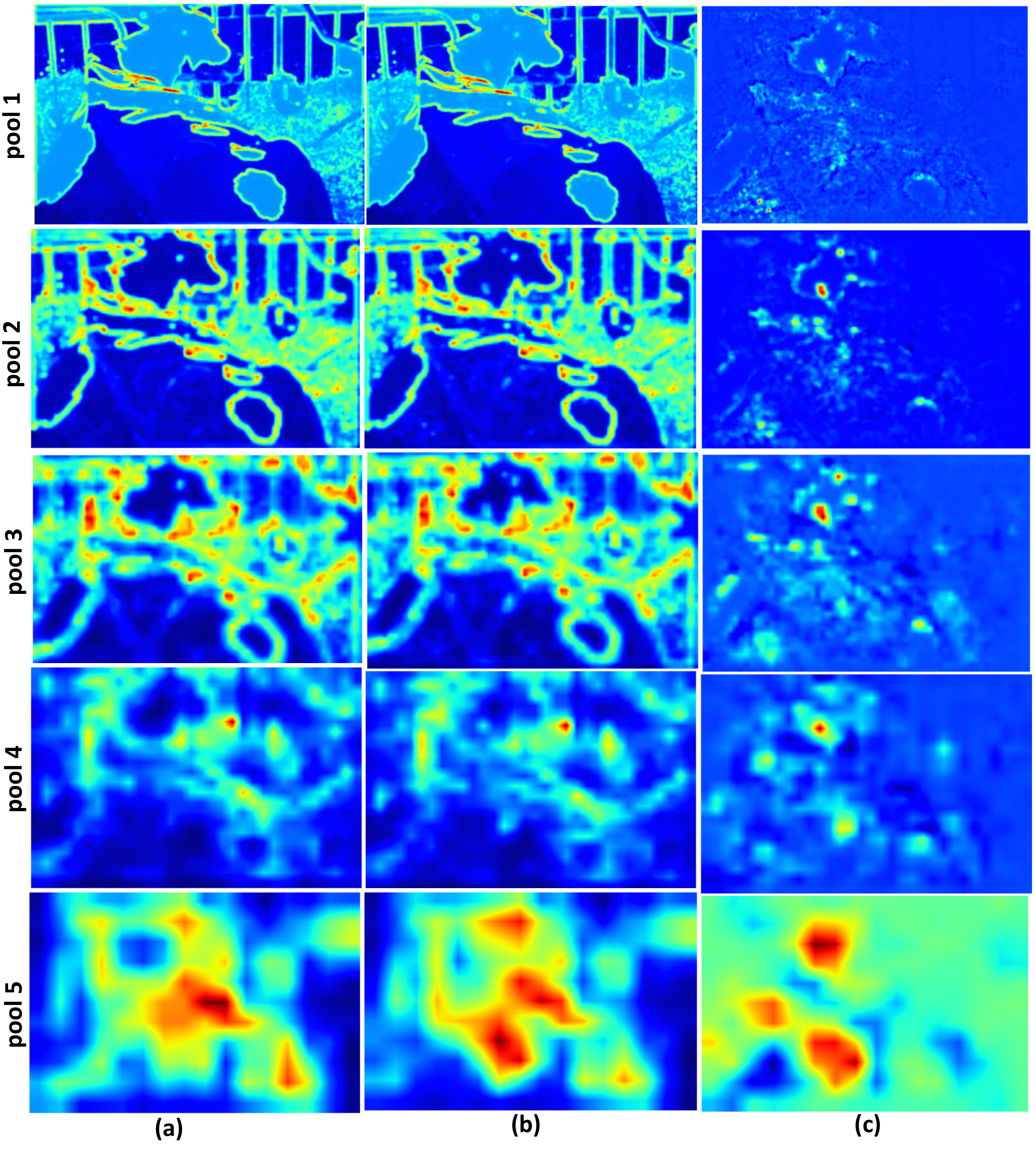} 
\caption{Visualization of filter responses showing how the correct context is propagated along the FCN-32s network. Column (a): filter responses during the forward pass using the original input. Column (b): filter responses during the forward pass using the perturbed input. Column (c): difference between (a) and (b)}
\label{fig:bigplot}
\vspace{-1mm}
\end{figure}

\begin{figure}[!b]
\vspace{-2mm}
\centering
\includegraphics[width=0.5\textwidth,height=2cm]{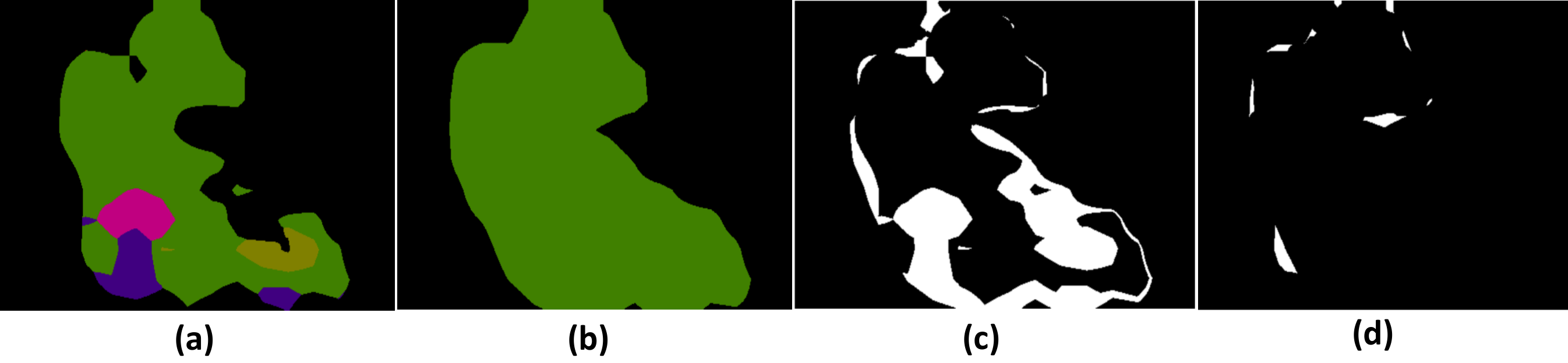} 
\caption{(a) Output of FCN-32s network (b) Output from the proposed approach (c) Pixels that were incorrectly classified by FCN-32s corrected by our approach (d) Pixels that were incorrectly classified by our approach that FCN-32s classified correctly.}
\label{fig:improved}
\vspace{-5mm}
\end{figure} 

\vspace{-5mm}
\subsubsection{Understanding Guided Perturbations}
\vspace{-1mm}
In this section, we perform several experiments to provide insight into different aspects of guided perturbations. Please refer to Figure~\ref{fig:approach} for the steps (Step 1, Step 2, Step 3)  mentioned in this section.

\textbf{Impact of perturbations on filter responses:} To get a clear understanding of what happens during the forward pass in Step 3 that vastly changes the network's prediction, we visualize the filter responses for the FCN-32s network in Figure \ref{fig:bigplot}. This model was chosen due to its simpler architecture but we observed similar behavior in other deep architectures too. In Figure \ref{fig:bigplot}, we plot the average filter responses at different layers through the deep network after upsampling them bilinearly to image size. As can be observed, the influence of the added perturbations are not visually explicit until the \textit{pool5} layer but the difference of the filter responses in Column (c) indicate that the information propagates from layers as early as \textit{pool2}.

Next, we analyze the pixels for which the network predictions changed from Step 1 to Step 3. Figure \ref{fig:improved}(c) shows the pixels that were classified wrongly during the forward pass in Step 1 but were correctly classified at the final output. On the other hand, \ref{fig:improved}(d) shows the pixels that were correctly classified in Step 1 but were incorrect at the final output. Observe that, the correctly classified pixels between Step 1 and Step 3 are mostly internal to the image where additional contextual information is available for the network to switch its prediction whereas the small number of misclassified pixels are largely concentrated along the boundary regions of the image where the context is ambiguous. We present more of such visualization examples in the supplementary material.

\begin{figure}[!htb]
\vspace{-2mm}
\centering
\includegraphics[width=0.5\textwidth,height=3.3cm]{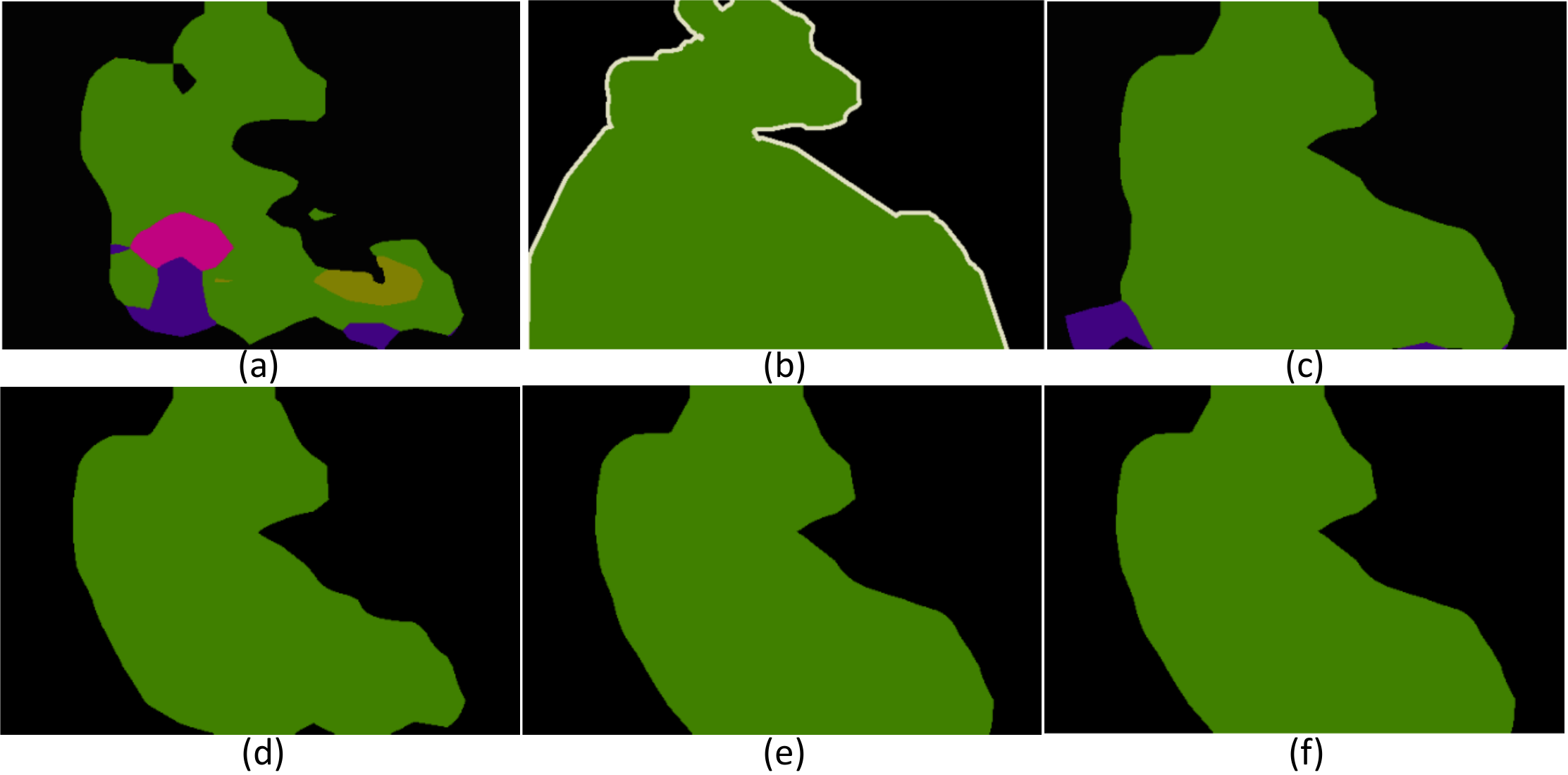} 
\vspace{-5mm}
\caption{(a) Output of FCN-32s (b)Ground truth labeling (c) Output of perturbed input using ground truth gradients (d)-(f) Output of perturbed input using guided perturbations  for iteration 1, 2 and 3 respectively.}
\label{fig:gt_gradient}
\vspace{-4mm}
\end{figure}

\textbf{Approximating ideal gradient direction:} In this experiment, we would like to answer the question: what are the ideal perturbations that can be generated at the input? The best one can do is to use the ground truth to generate error gradients at the softmax layer which is then backpropagated to generate the perturbed input. When this perturbed input is fed back to the network, the result is a vastly improved prediction as shown in Figure \ref{fig:gt_gradient} (c). 
While perturbations from ground truth significantly improve the performance, this information is not available during prediction time. The novelty of the current work is that the ground truth gradient direction is being approximated well enough by the predicted gradient directions that are computed using only the network's prediction.

To understand the extent of usefulness of the predicted gradients, we performed an experiment where the three steps outlined in our approach (Figure \ref{fig:approach}) is applied over successive iterations 
Figure \ref{fig:gt_gradient} (d) shows the output of our approach obtained in the first iteration and Figure \ref{fig:gt_gradient} (e)-(f) show the output over successive iterations. This shows that the most significant improvement happens at the first iteration and the subsequent iterations yield little improvement. We observe similar behavior on average over the PASCAL VOC2012 validation set.


\textbf{Intuition based on overlapping receptive fields:} In a CNN, the receptive fields of neighboring pixels define a context for their interactions. The advantage of having overlapping receptive fields is that the neighborhood connectivity is established automatically without explicitly specifying it. As long as the errors made by the CNN are sparse with respect to each pixel's receptive fields, the error gradients when accumulated over the entire network and used to perturb the input image exhibit a corrective behavior. The effect of GP can be seen as a type of residual information that is propagated through the network which results in contextual smoothing. This is evident by looking at the filter responses in Figure~\ref{fig:bigplot}, more specifically in Column 3, which shows the difference in responses with and without GP. It can be observed from the  \textit{pool5} responses that the peak activations occur around neighborhoods where there are competing classes. GP perturb these neighborhoods the most thus resulting in contextually smooth predictions in those regions. 

\begin{figure*}
\centering
    \subfloat[\label{fig:a}]{{\includegraphics[width=0.55\textwidth,height=0.3\linewidth]{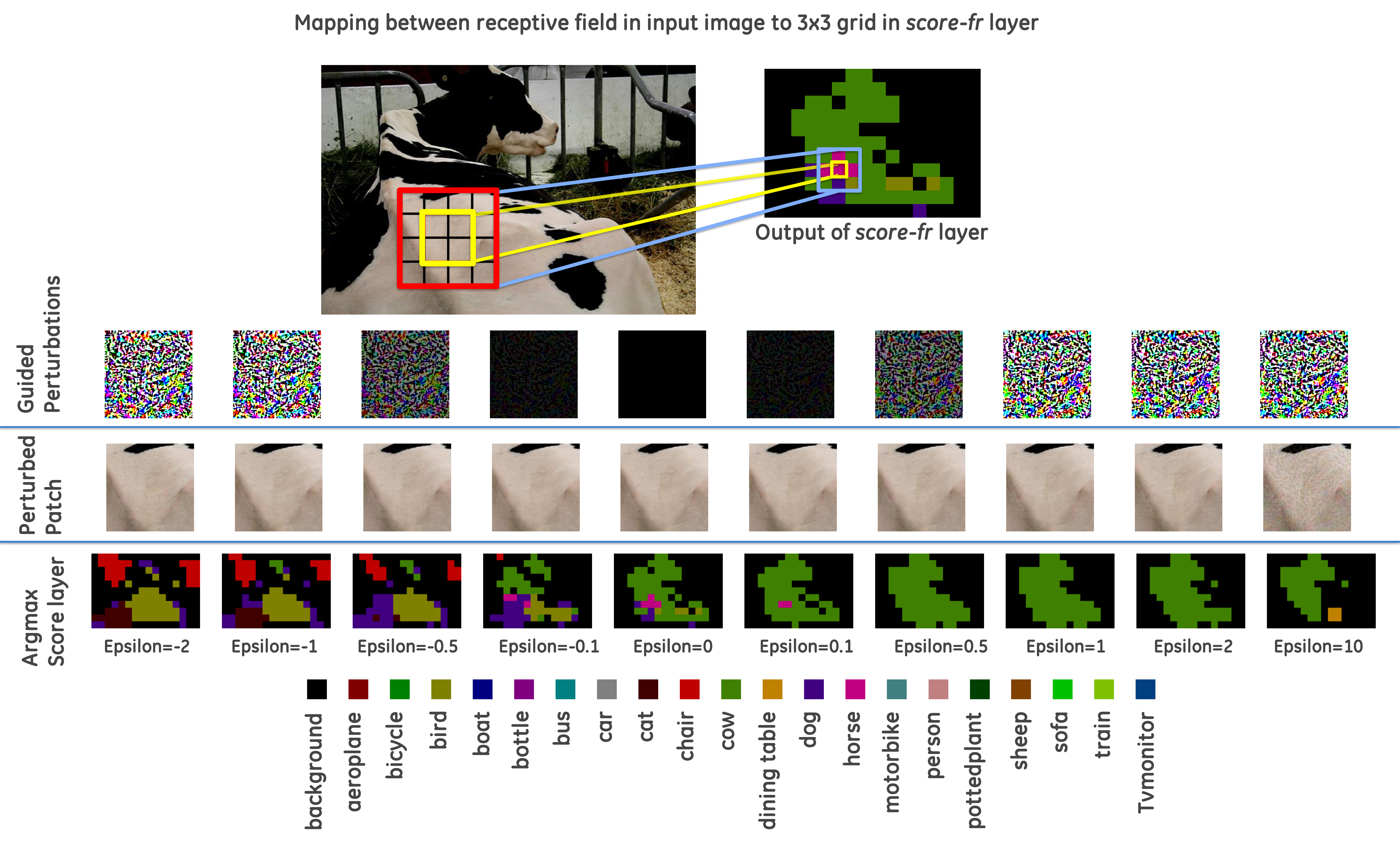}}}%
    \hfill
     \subfloat[\label{fig:c}]{{\includegraphics[width=0.4\textwidth,height=0.3\linewidth]{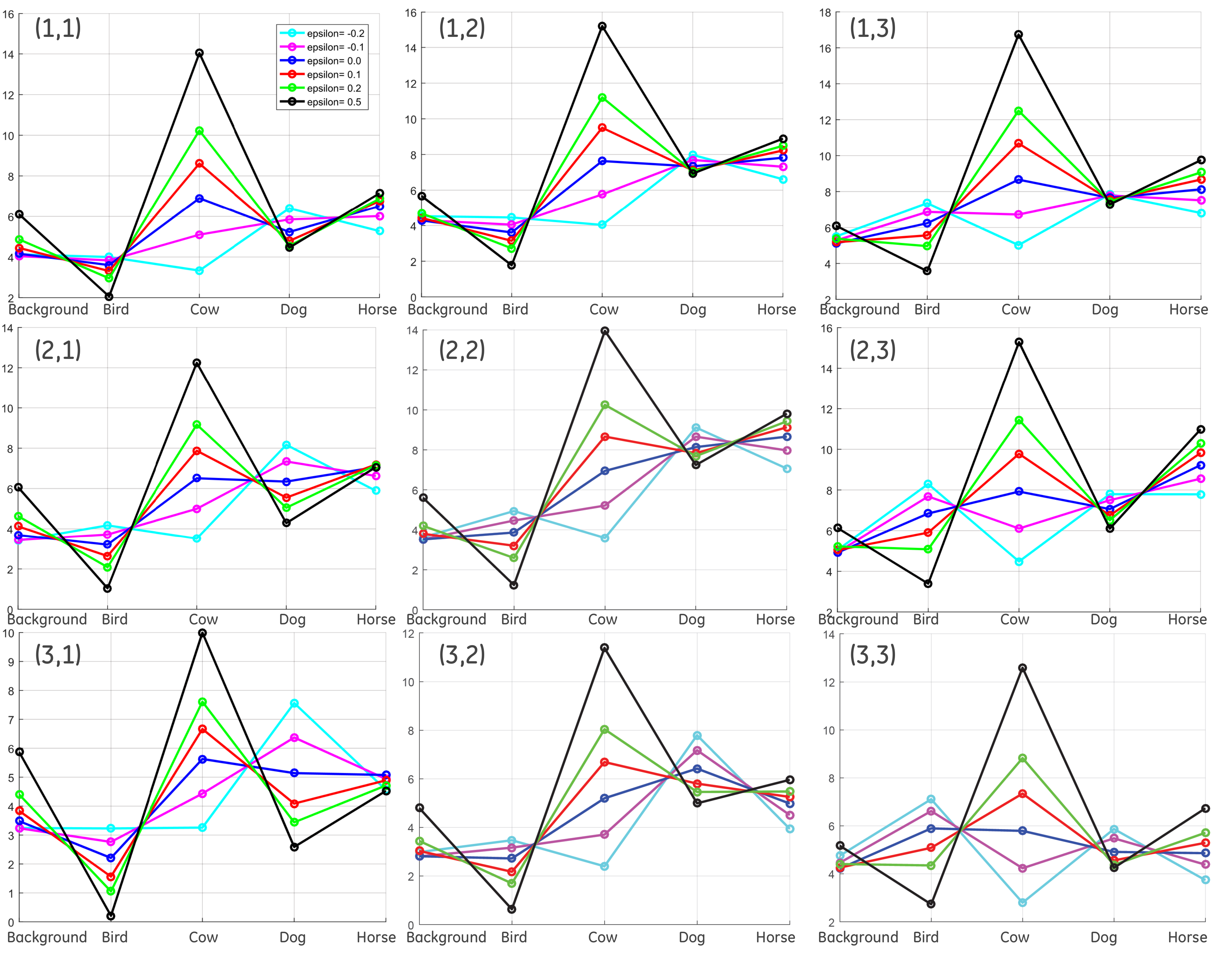}}}%
     \vspace{-3mm}
    \caption{(a) The top half shows an RGB patch in the input image and its corresponding patch in the \textit{score-fr} layer output of the FCN-32s network, before upsampling to image size. In the bottom half, we show, for different values of $\epsilon$ the guided perturbations, the perturbed RGB patches  and the \textit{score-fr} output. Notice how the the scores become contextually smoother for $\epsilon>0$. (b) The actual score values of the top-5 predicted classes for the 3x3 grid marked in blue in figure (a) are plotted. Observe that for a range of positive values of $\epsilon$, the correct class score (\textit{cow}) dominates the others across the entire neighborhood. The legend in (1,1) applies to all the plots. Best viewed in screen. Please zoom for clarity.}%
\label{fig:rfa}%
\vspace{-4mm}
\end{figure*}

\textbf{Analyzing GP in depth:} In figure~\ref{fig:rfa}, we show how guided perturbations impact the decisions made by the deep network by considering a local region in the input image and tracking its classification scores at \textit{score-fr} layer (before upsampling layer) across different values of $\epsilon$. In the top half of figure \ref{fig:a}, the patch of interest in the RGB image is marked by a red box and its corresponding region in the \textit{score-fr} output is marked by a blue box. Immediately below, the following are shown for different values of $\epsilon$: (1) Guided perturbations generated at the input (2) perturbed RGB patches (3) output of the \textit{score-fr} layer. Important observations that can be made from figure~\ref{fig:a} are:
\begin{itemize}
\vspace{-2mm}
\item Input perturbations corresponding to positive $\epsilon$ improve the score output over a vast range of values. This visualization shows how guided perturbations are able to operate at a local level by leveraging neighborhood contextual information as can be directly observed from the images of score layer shown in the bottom row.
\vspace{-2mm}
 \item Even a small negative value of $\epsilon$ results in a large adverse effect on the score output, without any perceptible change in the perturbed RGB patch. This shows that a negative $\epsilon$ corresponds to an adversarial setting.
\vspace{-1mm}
\end{itemize}

This discussion motivates our choice of using $\epsilon >0$ to generate GP. To further analyze how these input perturbations affect the actual classifier score, we show in figure~\ref{fig:c}, the predictions of the deep network for the 3x3 grid in the \textit{score-fr} output from figure~\ref{fig:a}, for different values of $\epsilon$. For clarity, we only show the predicted scores of the top-5 classes. From the score values of the grid position (2,2), we can observe that as $\epsilon$ increases, the score of true class (\textit{cow}) keeps increasing while the scores of the confusing classes do not vary much. The other plots show that this trend is observed across the entire neighborhood of the 3x3 grid. Thus, it can be inferred that perturbations at the input affect the decision of the deep network in a contextually consistent manner. We again observe that the score of the true class drops significantly even for a small negative $\epsilon$ which is consistent with our earlier observations. 

\begin{figure}[!h]

\subfloat{\includegraphics[width=0.5\textwidth]{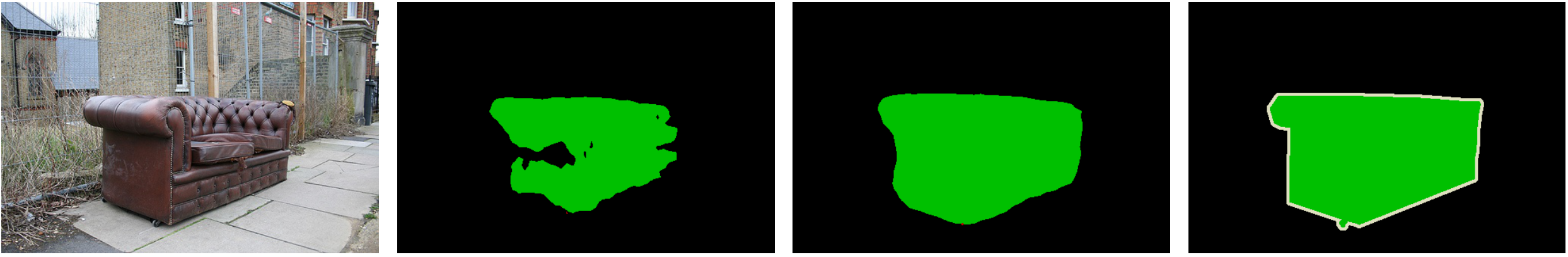}\label{c}}
\vspace{-3mm}
\newline
\subfloat{\includegraphics[width=0.5\textwidth]{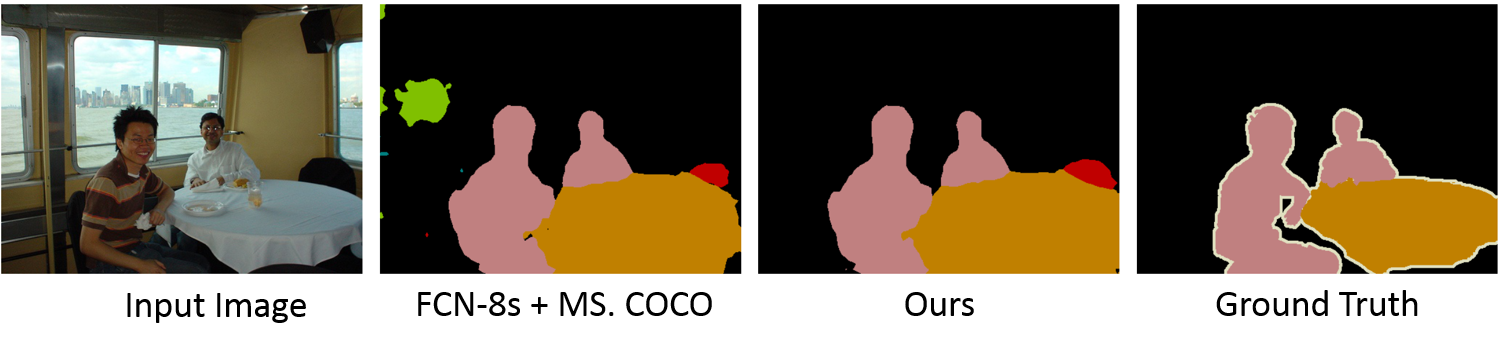}\label{d}}
\vspace{-4mm}
\newline
\subfloat{\includegraphics[width=0.5\textwidth]{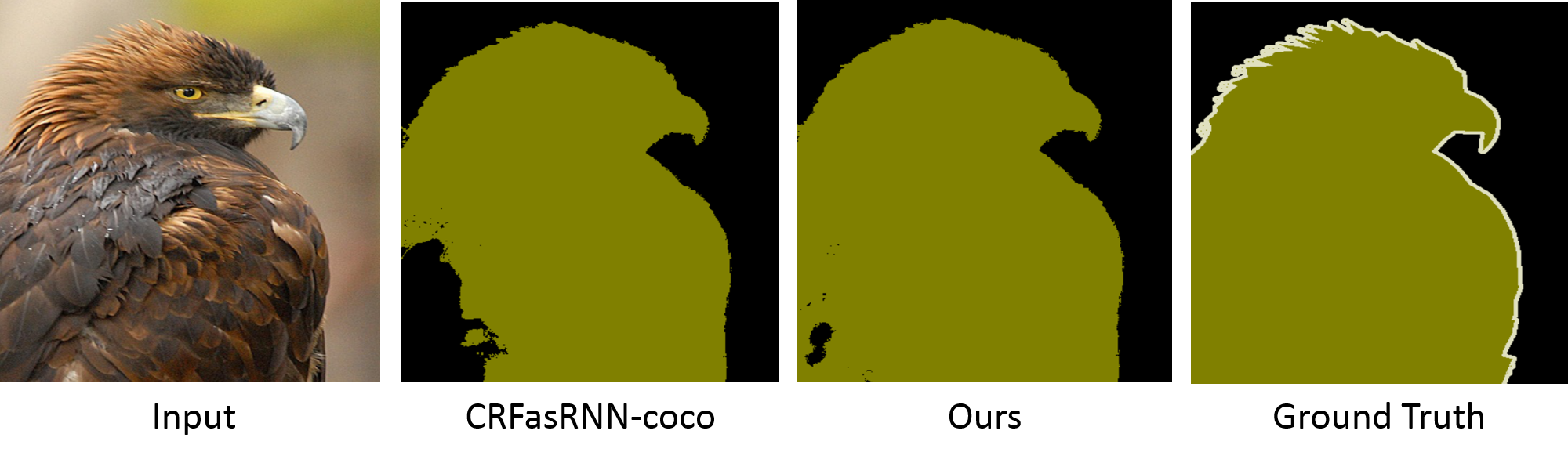}\label{f}}

\caption{Qualitative results on the PASCAL VOC2012 reduced validation set. In the top two rows, we compare our result with the FCN-8s part of CRFasRNN that has been trained on MS.COCO dataset \cite{mscocodb} and publicly released by \cite{crfasrnn}. In the bottom row, we compare with the complete CRFasRNN framework\cite{crfasrnn}. More results can be found in supplementary material.}
\label{figure9}
\vspace{-5mm}
\end{figure}

\section{Experiments}\label{sec:exp}
\vspace{-0.5mm}
In this section, we perform several experiments showing how our approach could be seamlessly applied on top of several pretrained deep networks. We test our method on the semantic segmentation task on PASCAL VOC2012 dataset \cite{pascalvocdb}, scene labeling task on the PASCAL Context 59-class dataset \cite{pascalcontextdb} and classification tasks on the MNIST and CIFAR10 datasets \cite{cifar10db}.
These results support how our approach is able to generalize across different types of problems in computer vision and highlights the advantage that it can be used with any pretrained model. 

\subsection{Evaluation Metrics}
\vspace{-0.5mm}
We evaluate our approach using the mean Intersection over Union (mIoU) metric commonly used for semantic segmentation as reported in \cite{fcncvpr}. Let $n_{ij}$ be the number of pixels of class $i$ predicted to belong to class $j$, $N_{cl}$ be number of classes, and $t_i = \sum_j n_{ij}$ be the total number of pixels of class $i$. It is then formulated as mean IoU = $\frac{1}{N_{cl}}\sum_i{\frac{n_{ii}}{t_i + \sum_j n_{ji} - n_{ii}}}$. For MNIST and CIFAR-10, we use classification accuracy as a metric to compare against the baseline. 

\subsection{Semantic Segmentation}\label{subsec:segresults}
\vspace{-0.5mm}
We use PASCAL VOC2012 dataset for evaluating our approach for semantic segmentation task. It consists of 21 classes including background. We use the following pre-trained models as baselines and show the improvement that can be obtained using our approach for each of them:(1) FCN-32s and FCN-8s \cite{fcncvpr}: these models are trained using the SBD dataset\cite{sbd} that consists of 9,600 images. (2) FCN-8s-coco and CRFasRNN \cite{crfasrnn}: these are trained using the images from MS COCO\cite{coco} and the SBD dataset using a total of 77,784 images. (3) Deeplab \cite{Deeplab16}: We evaluate on the Deeplab-VGG16 and ResNet101 models which use \textit{atrous} convolutions and multi scale evaluation. They are also trained on MS COCO and the SBD datasets.
\begin{table}[!t]
\caption{Results on the reduced VOC2012 validation set with 346 images. '-coco' denotes that the model was trained on MS COCO data in addition to the SBD dataset. Numbers in brackets show the magnitude of change compared to the corresponding base models.}
\label{tab: sem_seg}
\vspace{-2mm}
\small
\centering
\begin{tabular}{lll}
\hline
Method & \text{Base} & \text{Base+GP} \\ \hline
FCN-32s & 62.10 & 64.71 (\textbf{+2.6}) \\
FCN-8s & 63.97 & 66.97 (\textbf{+3.0}) \\ \hline
 & MS-COCO data & \\ \hline
FCN-8s-coco & 69.85 & 71.99 (\textbf{+2.1}) \\ 
CRFasRNN-coco & 72.95 & 73.75 (\textbf{+0.8}) \\ 
Deeplab-VGG16 & 66.9 & 69.1 (\textbf{+2.2}) \\ 
Deeplab-ResNet101 & 74.1 & 75.3 (\textbf{+1.2}) \\ \hline
\end{tabular}
\vspace{-4.5mm}
\end{table}

For all these methods, we use the publicly available models at the time of submission.  We use a single NVIDIA TitanX GPU for our experiments and CAFFE library\cite{caffe} for implementation. The pretrained models used in this section are obtained from the CAFFE Model Zoo~\cite{modelzoo} at the time of submission. All the reported results are computed with 1 iteration of our approach unless mentioned otherwise. Table \ref{tab: sem_seg} shows the results of applying the proposed approach to the different pretrained models during prediction time over a reduced validation set of 346 images as done in \cite{crfasrnn}.  As can be observed, the proposed approach results in increased performance over all the listed pretrained models. This reiterates the fact that our approach is indeed architecture independent and can be easily integrated even with complex feedforward architectures like CRFasRNN.  Table \ref{tab:voctest} shows the evaluation of our approach on PASCAL VOC2012 \textit{test set} using FCN-8s pretrained network as the base model to demonstrate the improvement shown by our method in an unbiased setting. The $\epsilon$ value used for the test set was tuned on the validation set.

\subsection{Scene Labeling}
The scene labeling task is a dense pixel labeling task that is evaluated on the PASCAL Context dataset. While there are more than 400 classes defined, the challenge entails evaluating on the 59 classes that are specified as most frequent \cite{pascalcontextdb}. The labeled classes contain scene elements in addition to objects that appear in the PASCAL VOC segmentation challenge, making this a much harder benchmark. To evaluate our approach on this task, we use the FCN-8s model from \cite{fcncvpr} as our baseline that was trained on the standard training split of 10,000 images provided with the dataset. The results, which were generated on the validation set consisting of 5105 images are shown in Table \ref{tab: scene_label}. We improve the performance of the FCN-8s network by 1.3\% which is significant given the large size of the validation set. Please note that the $\epsilon$ value was not tuned to fit this dataset rather the best performing $\epsilon$ from Table~\ref{tab: sem_seg} was used. 
\begin{table}[h!]
\caption{Results on the PASCAL-Context 59-classes validation set.}
\label{tab: scene_label}
\centering
\vspace{-5mm}
\begin{tabular}{l*{2}{>{$}c<{$}}}
\hline
Method & \text{mean IU} \\   
\hline
 FCN-8s & 39.12 \\ 
 FCN-8s + GP & \textbf{40.44} \\ 
 \end{tabular}
 \vspace{-4mm}
\end{table}

\vspace{-1.5mm}
\begin{table*}[!t]
\fontsize{14}{14}\selectfont
\caption{Results on the PASCAL VOC2012 test set consisting of 1456 images using FCN-8s as the base network. Use of Guided Perturbations improves the performance of the base network on 19 out of 21 classes.}
\label{tab:voctest}
\begin{center}
\vspace{-5mm}
\resizebox{\textwidth}{!}{\begin{tabular}{|*{22}{c|}| c ||}
    \hline
    \rule{0pt}{2ex}
    Method & bkg & aero & bicycle & bird & boat & bottle & bus & car & cat & chair & cow & table & dog & horse & mbk & person & plant & sheep & sofa & train & tv & mean \\ 
            \hline
     \rule{0pt}{2ex}
    FCN-8s \cite{fcncvpr} & 92.0 & 82.4 & \textbf{36.1} & 75.6 & 61.4 &	65.4 & 83.3	 & 77.2 &	80.1 &	27.9 &	66.8 &	51.5 &	73.6 &	71.9 &	78.9 &	77.1 &	55.3 &	73.4 &	44.3 &	74.0 &	\textbf{63.2} &	67.2	
    \\
    \hline
    \rule{0pt}{2ex}
     FCN-8s + GP & \textbf{92.4} & \textbf{84.4} & 35.9 & \textbf{79.3} & \textbf{62.6} & \textbf{70.5} & \textbf{86.2}	 & \textbf{80.0} &	\textbf{82.8} &	\textbf{28.0} &	\textbf{71.9} &	\textbf{55.2} &	\textbf{74.6} &	\textbf{75.6} &	\textbf{80.2} &	\textbf{77.4} &	\textbf{56.9} &	\textbf{75.6} &	\textbf{45.8} & \textbf{77.4} & 63.18 & \textbf{69.3}
    \\
    \hline
\end{tabular}}
\end{center}
\vspace{-5mm}
\end{table*}

\section{Ablative Experiments}\label{sec:ablative}
\vspace{-1mm}
For all the experiments in this section, we use FCN-32s network and the validation set used in section~\ref{subsec:segresults}.
\vspace{-4mm}
\paragraph{Speed-Performance trade-off}
The guided perturbations generated at the input layer of a deep network improves the performance of the base model. However, there is a computational overhead due to performing an additional backward and forward pass. As an alternative, the backward pass could be performed up to an intermediate layer in the deep network instead of the input layer. In this section, we provide results addressing the trade off between computational time and resulting performance due to perturbing layers other than the input. 
\begin{table}[h!]
\begin{center}
  \caption{Trade-off between performance and computation times obtained by truncating guided perturbations over different layers across the deep network. Original time taken is 0.12s per image. The baseline performance is 62.1\%}
\label{tab: tradeoff}
\vspace{-1.8mm}
    \begin{tabular}{ | l | l | l | l | l |}
    \hline
    layer & input & pool2 & pool3 & pool4 \\ \hline
    Time & 0.33s &  0.27s & 0.24s & 0.22s  \\ \hline    
    mIOU & 64.71 & 64.61 & 64.55 & 64.3 \\ \hline   
    \end{tabular}      
   \end{center}
   \vspace{-4.5mm}
\end{table}
It can be observed from Table \ref{tab: tradeoff} that even using the perturbed input from as late as \textit{pool4} layer the improvement in performance remains almost constant while computation time drops significantly. This experiment shows that effect of GP is not only observed at the input but also in the intermediate layers of the deep network and hence can be leverages for reducing the computational cost.
\paragraph{Comparison with CRF approaches} The behavior of GP resembles the contextual smoothing provided by VRF approaches that have been popularly used in Semantic Segmentation. In this section, we provide empirical evidence that GP captures additional dependencies in the data compared to pairwise interactions that are modeled by CRFs. Table \ref{tab: crf_results} shows the mean IoU values for cases with/without CRF applied on top of the network outputs. These results demonstrate that GP indeed captures extra dependencies compared to CRF and that GP can even improve upon CRF outputs.
\begin{table}[!t]
\vspace{-1mm}
\caption{Results with and without CRF}
\vspace{-2mm}
\label{tab: crf_results}
\small
\centering
\resizebox{\columnwidth}{!}{%
\begin{tabular}{ccccc}
\hline
Method & \text{Base} & \text{+CRF} & \text{+GP} & \text{+CRF+GP} \\ \hline
FCN-8s-coco & 69.8 & 71.1 & 72.0 & 72.7 \\
Deeplab-ResNet-101 & 74.1 & 74.9 & 75.3 & 75.8 \\ \hline
\end{tabular}
}
\end{table}

\vspace{-1mm}
\vspace{-1mm}
\paragraph{Guided Perturbations (vs) other strategies}\label{subsec:compareGP}
\vspace{-1mm}
In this section, we perform an ablative experiment where we perturb the input image in different ways in order to distinguish them from Guided Perturbations and show that the GP yields the most improvement in performance. As explained in Section~\ref{subsec:segment}, to generate a guided perturbation, we replace the softmax output with a one-hot encoded vector for the class of maximum confidence. We consider different methods to modify the label distribution that is obtained from the softmax function as follows: 

\begin{itemize}
\vspace{-2mm}
\itemsep0em
\item \textit{random-onehot}: The class label is chosen in an uniformly random manner and used as ground truth instead of the maximum probability class.
\vspace{-1mm}
\item \textit{Uniform-label}: An uniform label distribution is produced by assigning equal probability to all the classes and used as encoding to generate the error gradient.
\vspace{-1mm}
\item \textit{top2-label}: Modified label distribution contains equal probability to top two predicted classes and used as encoding to generate the error gradient.
\end{itemize}
\vspace{-2mm}

\begin{figure}[!h]
\vspace{-1mm}
\centering
\includegraphics[width=0.45\textwidth]{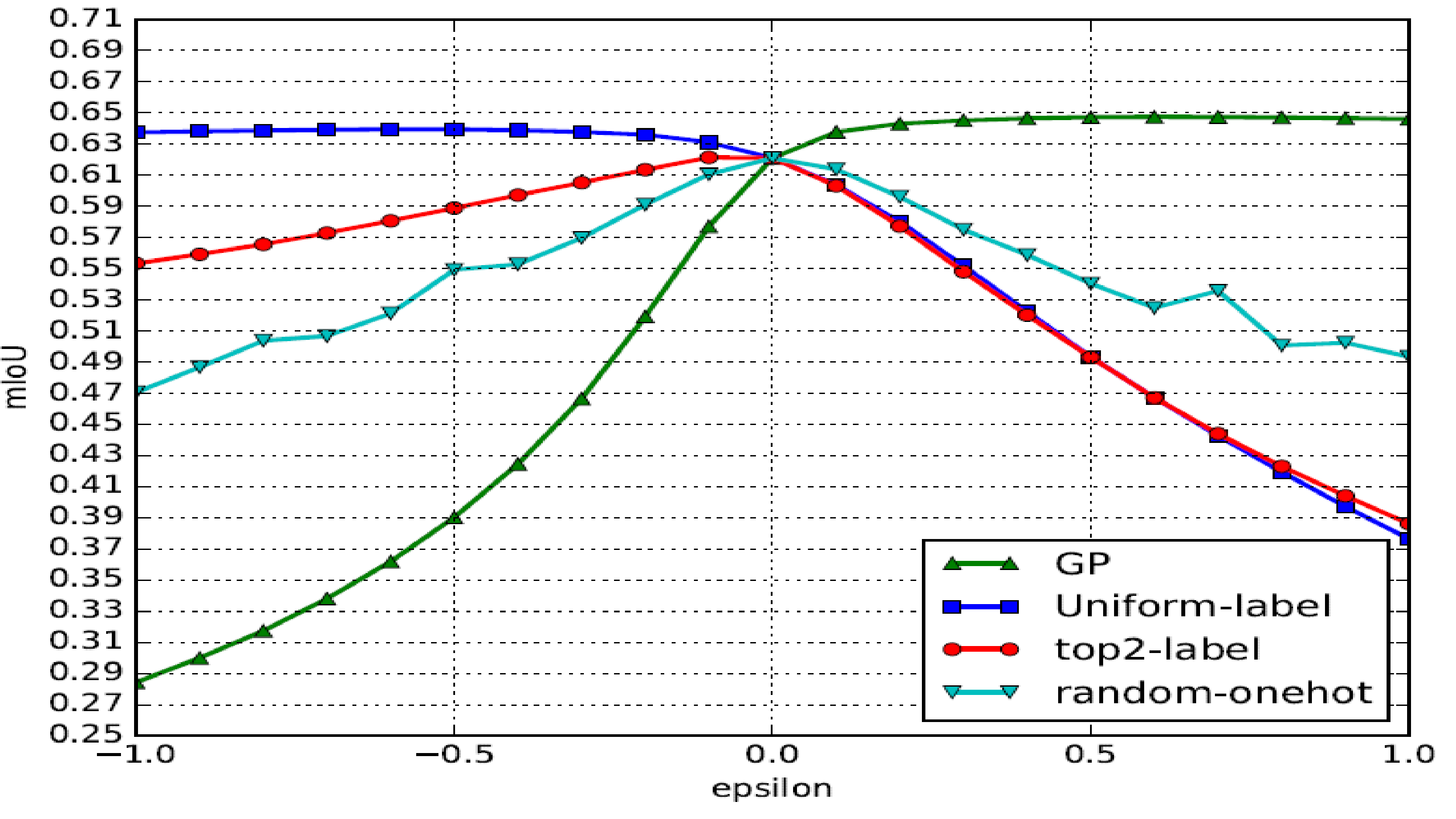} 
\vspace{-2mm}
\caption{Mean IOU values for several perturbations generated by using different types of label distributions on the validation set over the range $\epsilon=[-1,1]$ with FCN-32s as the base network. Please refer to section~\ref{subsec:compareGP} for details.}
\label{fig:gpvsrest}
\vspace{-3mm}
\end{figure}

Figure~\ref{fig:gpvsrest} shows the effect of different types of label distribution on the segmentation performance. At the outset, it can be observed that GP gives the best quantitative performance of $64.7\%$ compared to the second best case, which is the uniform setting with negative $\epsilon$ which scores $63.8\%$. We can also observe that when we perform GP, $\epsilon < 0$ corresponds to the adversarial setting. Intuitively, this setting is equivalent to maximizing the loss of the softmax classification function during training. Hence, the backpropagated gradient always moves away from the correct class. In our approach, GP is always generated by setting $\epsilon > 0$ as mentioned in sections~\ref{subsec:segment} and ~\ref{subsec:class}. The setting involving choosing a random label to generate the one-hot vector at the softmax output results in poor performance across all values of $\epsilon$ since gradient directions become random and the resulting perturbations adversely affect the performance of the deep network on the perturbed input image. 

\begin{figure}[!h]
\vspace{-2mm}
\centering
\includegraphics[width=0.5\textwidth]{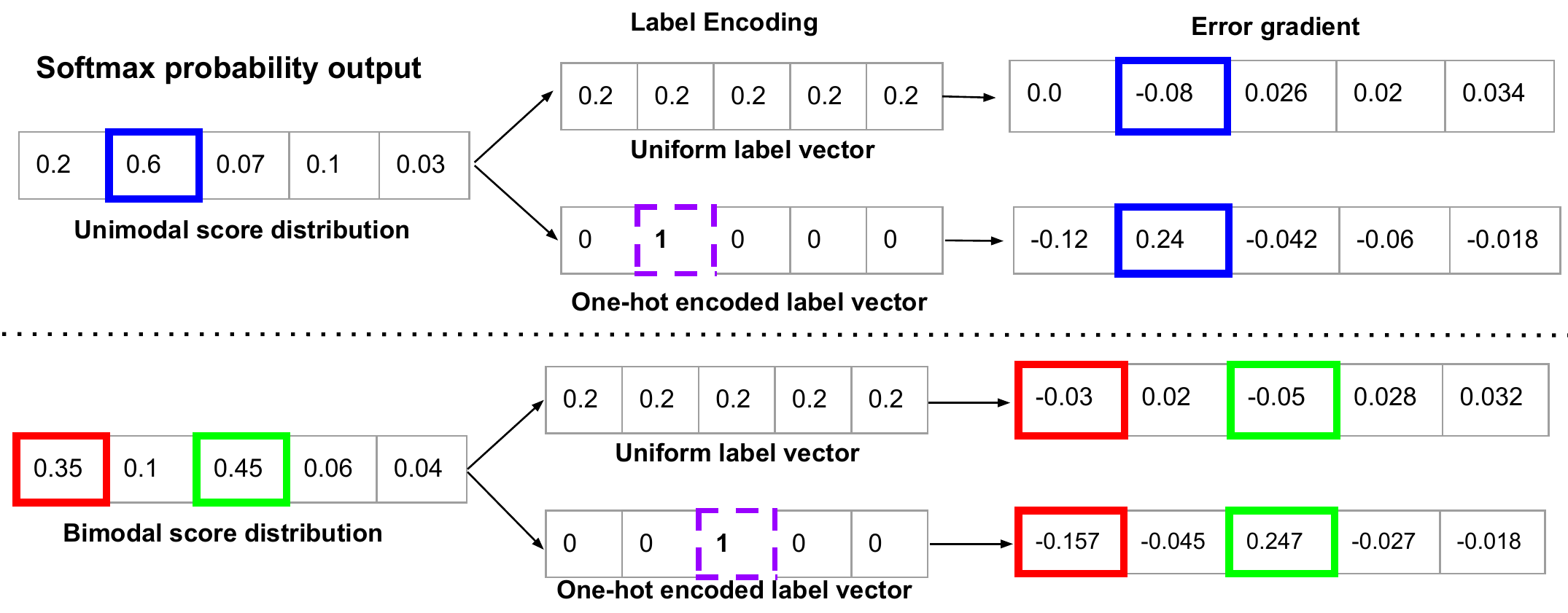} 
\caption{Difference in the gradient signal generated between \textit{Uniform-label} setting and GP for the case of unimodal output score distribution (top) and bimodal output score distribution (bottom). The dominant gradient direction in both cases is shown in the colored boxes. The exact derivation for computing these gradient values is given in the supplementary material.}
\label{fig:unifGP}
\vspace{-3mm}
\end{figure}

The interesting case to analyze from Figure~\ref{fig:gpvsrest} is the performance of the \textit{Uniform-label} setting for $\epsilon < 0$. 
To understand this effect, Figure~\ref{fig:unifGP} illustrates a toy example showing the difference between the error gradients generated using GP and \textit{Uniform-label} setting for a different possible output score distributions from the CNN. In this toy example, the CNN is trained to classify among 5 classes.  Observe that, for the unimodal case, the gradient signal generated for a uniform output label distribution has the same relative magnitude as the gradient signal generated for GP but the dominant gradient direction is exactly the opposite. However, GP still gives a better performance compared to the uniform label distribution. In this case, the score vector is bimodal and hence there are two dominant directions in the gradient signal. Notice that the top gradient direction in the case of GP still points towards the correct class and all other directions move away from the correct class, as expected. But in the case of uniform label distribution, there are two competing directions and hence there is higher probability for the gradient to move in the wrong direction. 

\begin{figure}[!h]
\includegraphics[width=0.5\textwidth,height=0.15\textwidth]{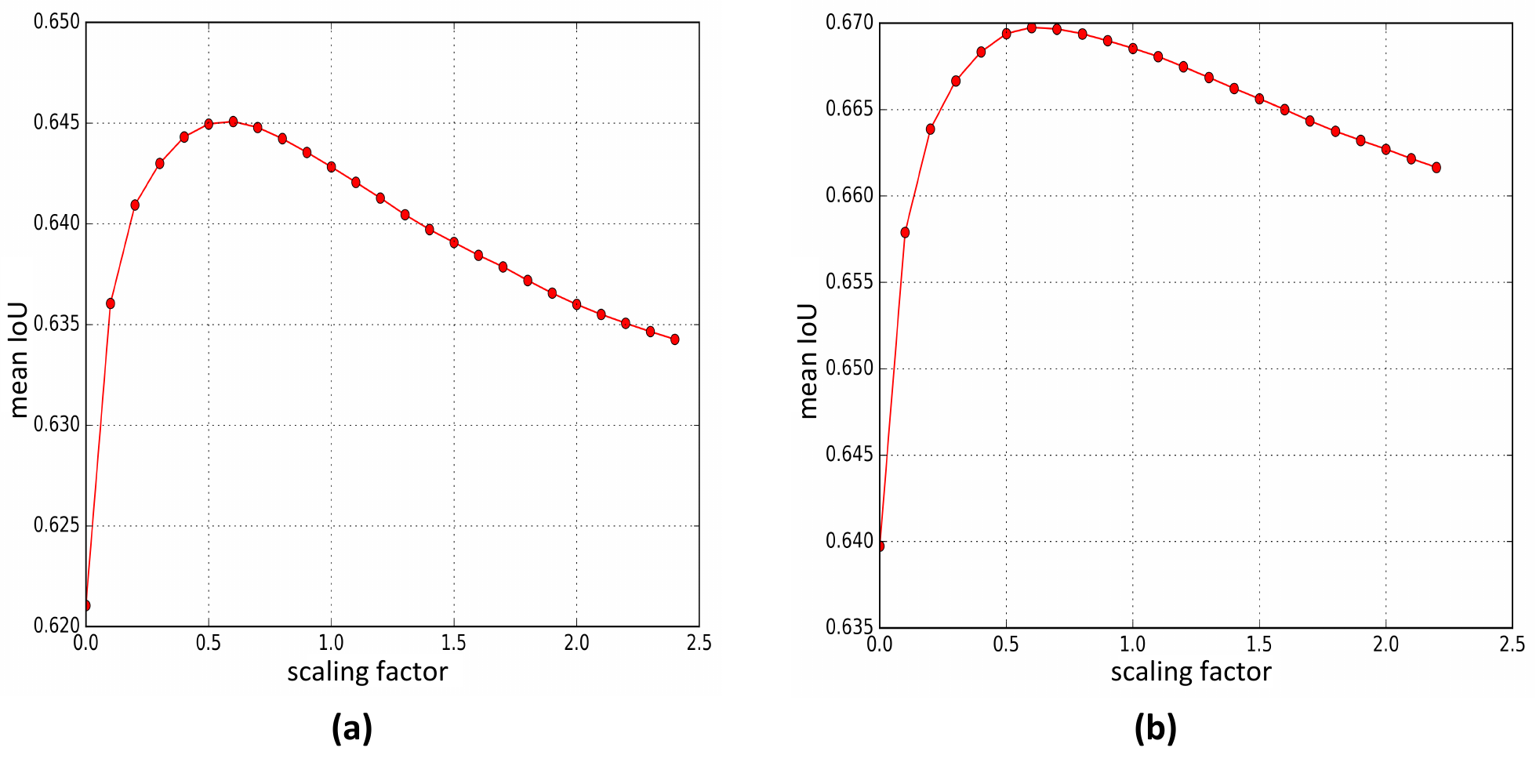}
\vspace{-8mm}
\caption{Effect of scaling factor $\epsilon$ on performance of FCN-32s (left) and FCN-8s (right) networks evaluated on the reduced PASCAL VOC2012 validation set. Best viewed in screen. Please zoom for clarity.}
\label{fig:scale}
\vspace{-6mm}
\end{figure}

\begin{figure}[!th]
\centering
\includegraphics[width=0.35\textwidth,height=0.1\textwidth]{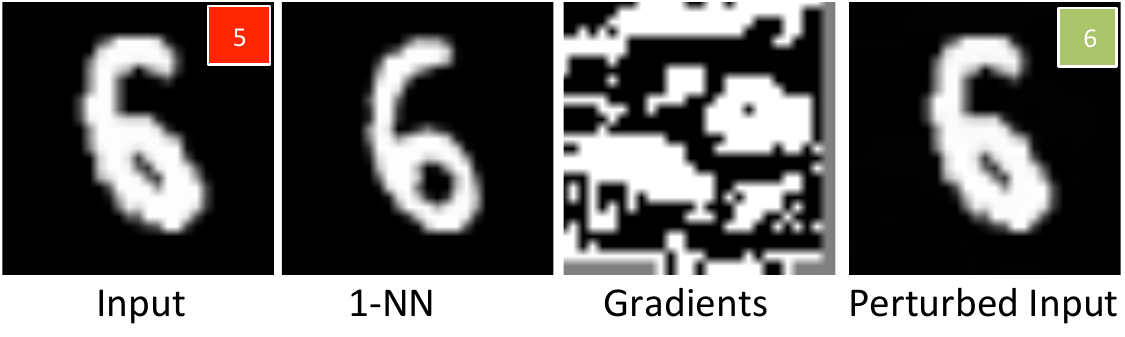} 
\vspace{-2mm}
\caption{The input image is classified as `5'. By perturbing the input from the gradients generated using the top nearest neighbor class, the network changes its prediction to `6'}
\vspace{-4mm}
\label{fig:cls}
\end{figure}

\vspace{-3mm}
\paragraph{Effect of Scaling parameter}\label{sec:scaling}
We evaluate the performance of our approach using FCN-32s and FCN-8s networks over a range of scaling parameter $\epsilon$ on the validation set. Figure \ref{fig:scale} shows how the performance varies based on the scaling factor. It can be observed that improvement in performance is generally obtained over a wide range of values of $\epsilon$. This indicates that network's behavior is not very sensitive to the value of $\epsilon$ though there seems to be an optimal value for best performance that depends on the deep model. We use $\epsilon=0.55$ for FCN-32s, $\epsilon=0.7$ for FCN-8s network and $\epsilon=0.22$ for CRFasRNN network for our experiments.  
\vspace{-2mm}
\paragraph{Image Classification}\label{subsec:class}
The method described in Section \ref{subsec:segment} for semantic segmentation cannot be applied directly for classification tasks. Since \textit{context} for a classification task is not defined naturally, we extract contextual information from the learned feature space.  Given an input image, we first extract the feature from the deep network and use it to select top $k$ nearest neighbors from the training set using euclidean distance metric. We then perturb the test image with the weighted average of gradients generated using the class of the selected nearest neighbors and perform a forward pass to predict the final output. Let ${nn_{i}}$ be the class of the $i^{th}$ nearest neighbor. Following the notation established in Section~\ref{subsec:segment}, the equation for perturbed image is given as follows: $\mathbb{X}_{per}=\mathbb{X}+\epsilon \sum^k_{i=1} (w_{i}sign(\nabla_{X} \mathcal{J}(\theta, \mathbb{X}_i,\mathbb{Y}_{nn_{i}})))$, where $\mathbb{X}_i$ is the $i^{th}$ nearest neighbor; $k$ is the number of nearest neighbors and $w_i$ is weight associated with each nearest neighbor $i$ and $\mathcal{J}(.)$ corresponds to the loss function. Figure \ref{fig:cls} shows an example where the network correctly classifies the perturbed input generated using this procedure.

\vspace{-1mm}
To evaluate the performance of GP on classification, we tested the method on two standard datasets: MNIST and CIFAR10. MNIST consists of grayscale images of digits while CIFAR10 consists of more realistic images of object classes. We follow the standard training/testing split for both the cases. We use 3 nearest neighbor with equal weights for all our experiments. For MNIST, we use a CNN with 2 conv. layers and 2 fully-connected layers with a 20-50-500-10 architecture and for CIFAR10, we use a CNN with 5 conv. and 2 fully-connected layers with a 64-64-128-128-128-128-10 architecture.

\begin{table}[htb]
\vspace{-1mm}
\caption{Results on the classification task on MNIST and CIFAR10 datasets.}
\label{tab: mnist_cifar}
\centering
\vspace{-3mm}
\begin{tabular}{l*{3}{>{$}c<{$}}}
\hline
Dataset & \text{Baseline} & \text{Proposed} \\   
\hline
MNIST  & 98.92 & \textbf{99.15} \\ 
CIFAR10 & 76.31 & \textbf{76.95} \\ 
 \end{tabular}
 \vspace{-4mm}
\end{table}

Table \ref{tab: mnist_cifar} shows the results of our classification experiments. GP improves performance over the baseline on both the datasets. However, the improvement in performance is not as high as in the segmentation case which could be attributed to two reasons: (1) the base networks themselves have learned a very strong representation and (2) the context information in the  classification task is relatively weak compared to the segmentation task.

\vspace{-3mm}
\section{Conclusion}\label{sec:conclude}
\vspace{-1mm}
In this paper, we have shown novel self-corrective behavior of CNNs for segmentation and classification tasks. We showed that guided perturbations can improve the network's performance without additional training or network modification. We have demonstrated this effect on several publicly available datasets and  using different network architectures. We have presented several experiments that try to understand and explain different aspects of guided perturbations. We believe that this behavior can lead to novel network designs and better end-to-end training procedures.
{\small
\bibliographystyle{ieee}
\bibliography{refs}

\begin{thebibliography}{10}\itemsep=-1pt

\bibitem{modelzoo}
Caffe model zoo.
\newblock \url{https://github.com/BVLC/caffe/wiki/Model-Zoo}.
\newblock Accessed: 2010-09-30.

\bibitem{chen2016deeplab}
L.-C. Chen, G.~Papandreou, I.~Kokkinos, K.~Murphy, and A.~L. Yuille.
\newblock Deeplab: Semantic image segmentation with deep convolutional nets,
  atrous convolution, and fully connected crfs.
\newblock {\em arXiv preprint arXiv:1606.00915}, 2016.

\bibitem{Deeplab16}
L.-C. Chen, G.~Papandreou, I.~Kokkinos, K.~Murphy, and A.~L. Yuille.
\newblock Deeplab: Semantic image segmentation with deep convolutional nets,
  atrous convolution, and fully connected crfs.
\newblock {\em arXiv:1606.00915}, 2016.

\bibitem{pascalvocdb}
M.~Everingham, S.~A. Eslami, L.~Van~Gool, C.~K. Williams, J.~Winn, and
  A.~Zisserman.
\newblock The pascal visual object classes challenge: A retrospective.
\newblock {\em International Journal of Computer Vision}, 111(1):98--136, 2015.

\bibitem{adv2014}
I.~J. Goodfellow, J.~Shlens, and C.~Szegedy.
\newblock Explaining and harnessing adversarial examples.
\newblock {\em arXiv preprint arXiv:1412.6572}, 2014.

\bibitem{sbd}
B.~Hariharan, P.~Arbel{\'a}ez, L.~Bourdev, S.~Maji, and J.~Malik.
\newblock Semantic contours from inverse detectors.
\newblock In {\em 2011 International Conference on Computer Vision}, pages
  991--998. IEEE, 2011.

\bibitem{caffe}
Y.~Jia, E.~Shelhamer, J.~Donahue, S.~Karayev, J.~Long, R.~Girshick,
  S.~Guadarrama, and T.~Darrell.
\newblock Caffe: Convolutional architecture for fast feature embedding.
\newblock In {\em Proceedings of the 22Nd ACM International Conference on
  Multimedia}, MM '14, pages 675--678. ACM, 2014.

\bibitem{cifar10db}
A.~Krizhevsky.
\newblock Learning multiple layers of features from tiny images.
\newblock Technical report, 2009.

\bibitem{coco}
T.~Lin, M.~Maire, S.~J. Belongie, L.~D. Bourdev, R.~B. Girshick, J.~Hays,
  P.~Perona, D.~Ramanan, P.~Doll{\'{a}}r, and C.~L. Zitnick.
\newblock Microsoft {COCO:} common objects in context.
\newblock {\em CoRR}, abs/1405.0312, 2014.

\bibitem{mscocodb}
T.-Y. Lin, M.~Maire, S.~Belongie, J.~Hays, P.~Perona, D.~Ramanan,
  P.~Doll{\'a}r, and C.~L. Zitnick.
\newblock Microsoft coco: Common objects in context.
\newblock In {\em European Conference on Computer Vision}, pages 740--755.
  Springer, 2014.

\bibitem{inversion2015}
A.~Mahendran and A.~Vedaldi.
\newblock Understanding deep image representations by inverting them.
\newblock In {\em 2015 IEEE conference on computer vision and pattern
  recognition (CVPR)}, pages 5188--5196. IEEE, 2015.

\bibitem{pascalcontextdb}
R.~Mottaghi, X.~Chen, X.~Liu, N.-G. Cho, S.-W. Lee, S.~Fidler, R.~Urtasun, and
  A.~Yuille.
\newblock The role of context for object detection and semantic segmentation in
  the wild.
\newblock In {\em Proceedings of the IEEE Conference on Computer Vision and
  Pattern Recognition}, pages 891--898, 2014.

\bibitem{nguyen2015deep}
A.~Nguyen, J.~Yosinski, and J.~Clune.
\newblock Deep neural networks are easily fooled: High confidence predictions
  for unrecognizable images.
\newblock In {\em 2015 IEEE Conference on Computer Vision and Pattern
  Recognition (CVPR)}, pages 427--436. IEEE, 2015.

\bibitem{ILSVRC15}
O.~Russakovsky, J.~Deng, H.~Su, J.~Krause, S.~Satheesh, S.~Ma, Z.~Huang,
  A.~Karpathy, A.~Khosla, M.~Bernstein, A.~C. Berg, and L.~Fei-Fei.
\newblock {ImageNet Large Scale Visual Recognition Challenge}.
\newblock {\em International Journal of Computer Vision (IJCV)},
  115(3):211--252, 2015.

\bibitem{fcncvpr}
E.~Shelhamer, J.~Long, and T.~Darrell.
\newblock Fully convolutional networks for semantic segmentation.
\newblock {\em IEEE Transactions on Pattern Analysis and Machine Intelligence},
  PP(99):1--1, 2016.

\bibitem{intriguing}
C.~Szegedy, W.~Zaremba, I.~Sutskever, J.~Bruna, D.~Erhan, I.~Goodfellow, and
  R.~Fergus.
\newblock Intriguing properties of neural networks.
\newblock {\em arXiv preprint arXiv:1312.6199}, 2013.

\bibitem{gcrf}
R.~Vemulapalli, O.~Tuzel, M.-Y. Liu, and R.~Chellappa.
\newblock Gaussian conditional random field network for semantic segmentation.
\newblock CVPR, 2016.

\bibitem{deepvis}
J.~Yosinski, J.~Clune, A.~Nguyen, T.~Fuchs, and H.~Lipson.
\newblock Understanding neural networks through deep visualization.
\newblock {\em arXiv preprint arXiv:1506.06579}, 2015.

\bibitem{mscontext}
F.~Yu and V.~Koltun.
\newblock Multi-scale context aggregation by dilated convolutions.
\newblock {\em arXiv preprint arXiv:1511.07122}, 2015.

\bibitem{deconv}
M.~D. Zeiler, D.~Krishnan, G.~W. Taylor, and R.~Fergus.
\newblock Deconvolutional networks.
\newblock In {\em Computer Vision and Pattern Recognition (CVPR), 2010 IEEE
  Conference on}, pages 2528--2535. IEEE, 2010.

\bibitem{crfasrnn}
S.~Zheng, S.~Jayasumana, B.~Romera-Paredes, V.~Vineet, Z.~Su, D.~Du, C.~Huang,
  and P.~H. Torr.
\newblock Conditional random fields as recurrent neural networks.
\newblock In {\em Proceedings of the IEEE International Conference on Computer
  Vision}, pages 1529--1537, 2015.

\end{thebibliography}
}
\onecolumn
\begin{center}
\section*{\LARGE{Appendix}}\label{sec:app}
\end{center}

\begin{figure}
\centering
\subfloat{\includegraphics[width=\textwidth,height=0.7\linewidth]{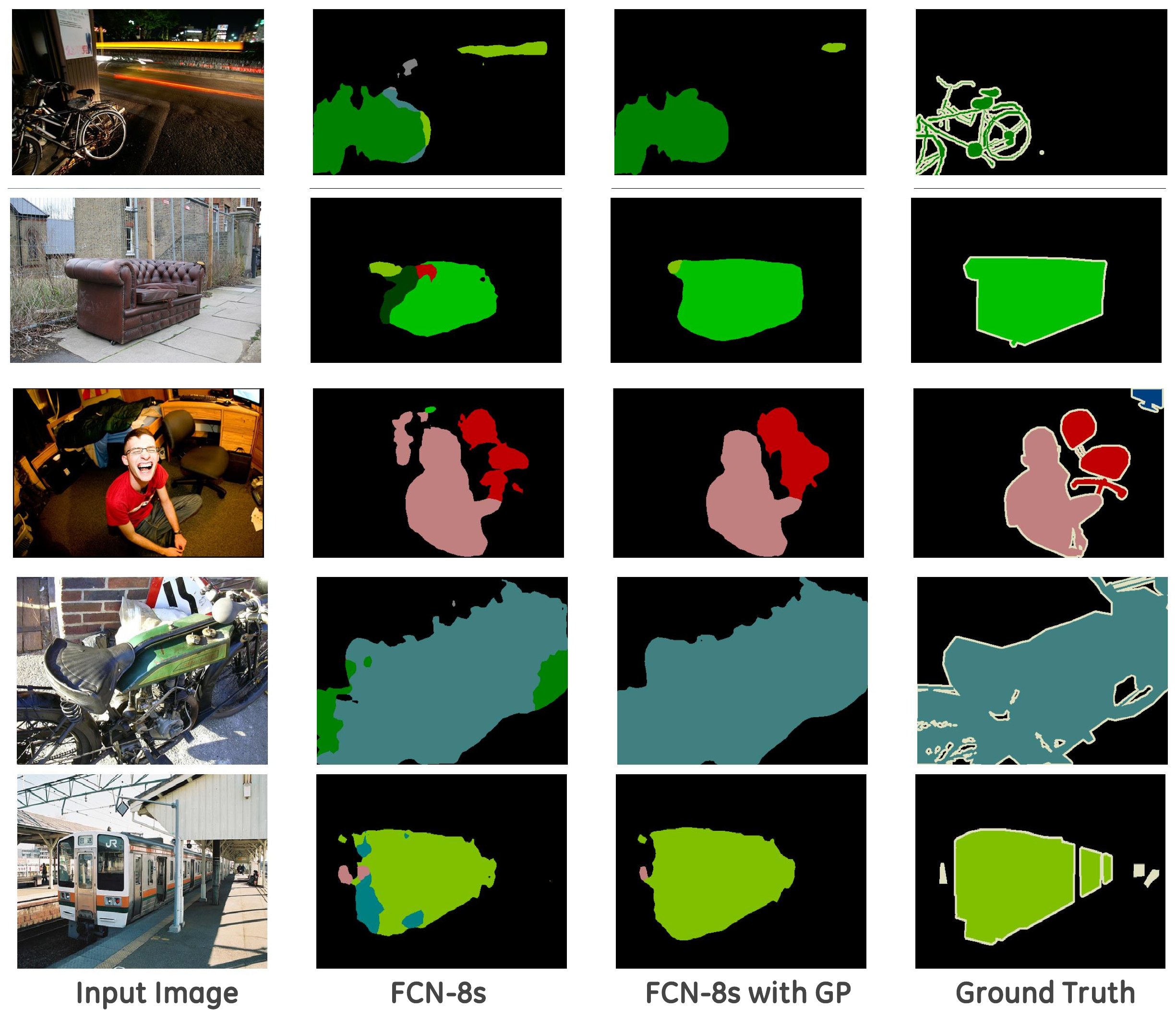}\label{c}}
\newline
\subfloat{\includegraphics[width=\textwidth,height=0.5\linewidth]{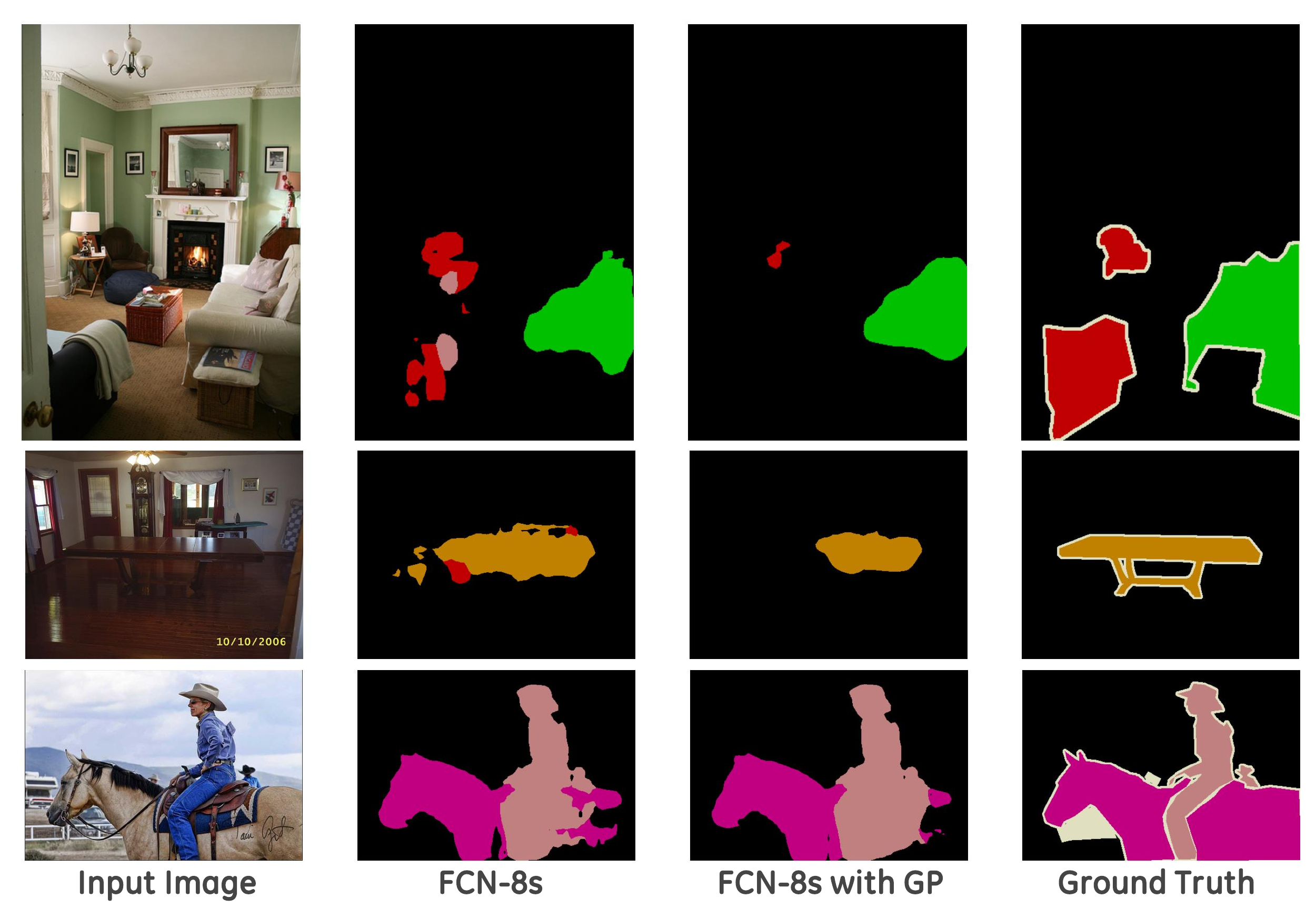}\label{d}}
\newline
\vspace{-5mm}
\caption{Qualitative results on the PASCAL VOC2012 reduced validation set - Comparison with FCN-8s pretrained model. Top half shows the successful outputs, Bottom half shows the failure cases.}
\label{fig:fcn8s_outputs_app}
\end{figure}

\begin{figure}
\centering
\subfloat{\includegraphics[width=\textwidth,height=0.7\linewidth]{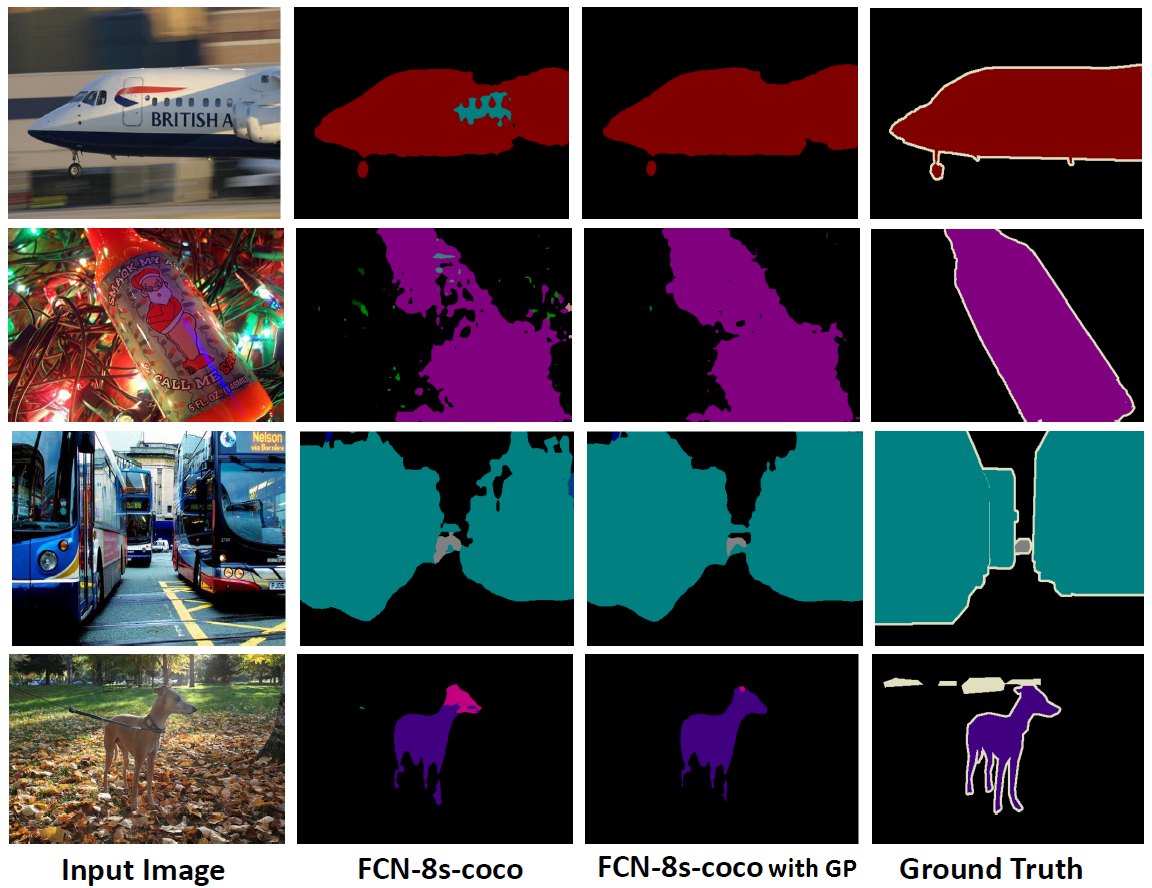}\label{c}}
\newline
\subfloat{\includegraphics[width=\textwidth,height=0.45\linewidth]{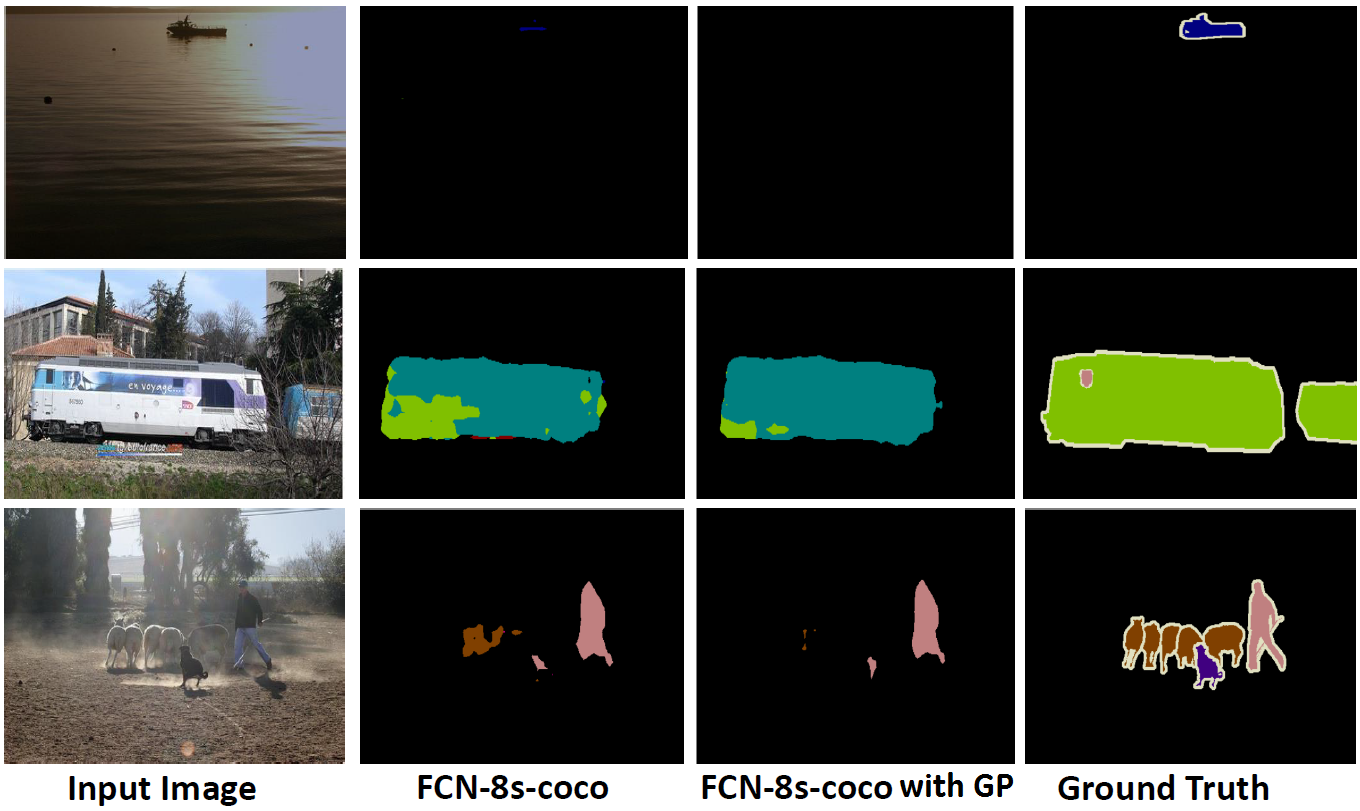}\label{d}}
\newline
\caption{Qualitative results on the PASCAL VOC2012 reduced validation set - Comparison with FCN-8s-coco pretrained model. Top half shows the successful outputs, Bottom half shows the failure cases.}
\label{fig:fcn8s_coco_outputs_app}
\end{figure}

\begin{figure}
\centering
\subfloat{\includegraphics[width=\textwidth,height=0.7\linewidth]{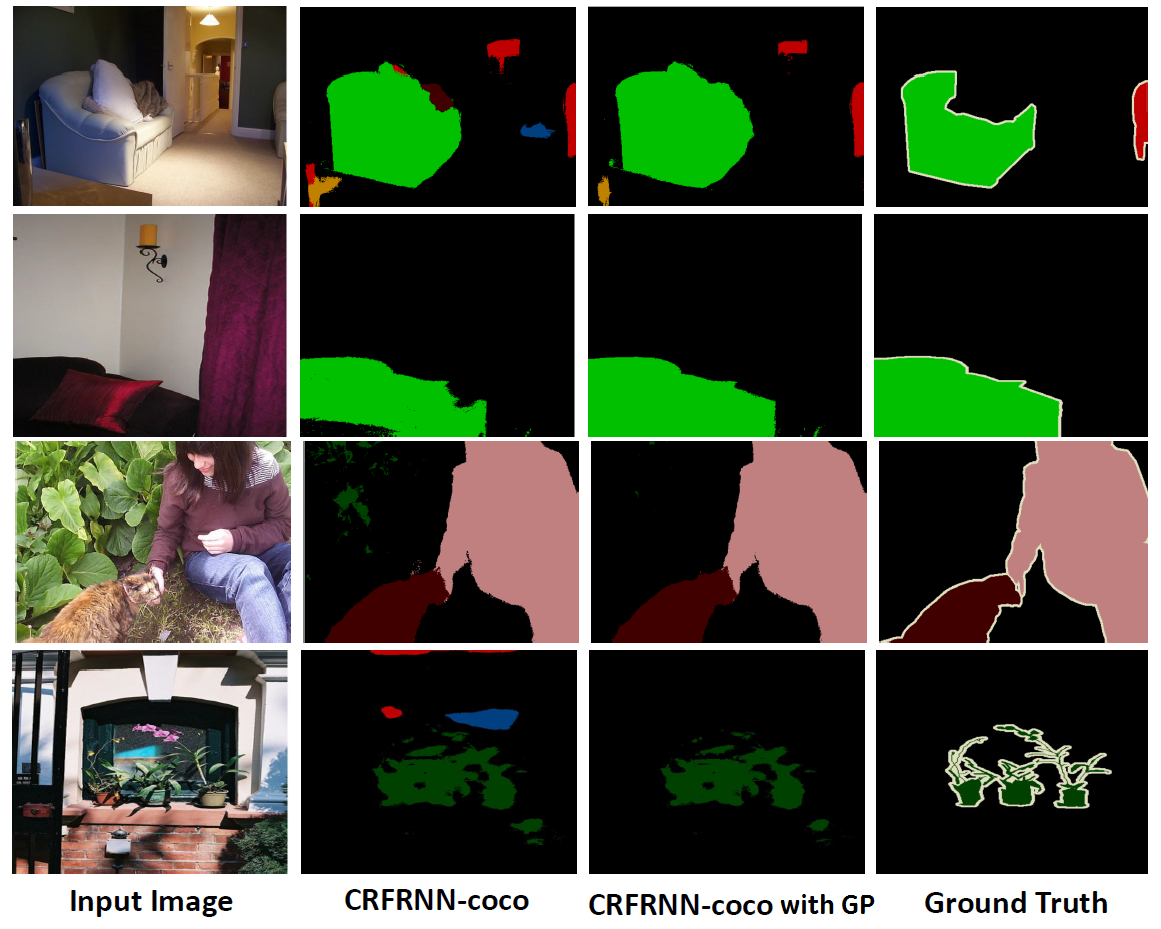}\label{c}}
\newline
\subfloat{\includegraphics[width=\textwidth,height=0.45\linewidth]{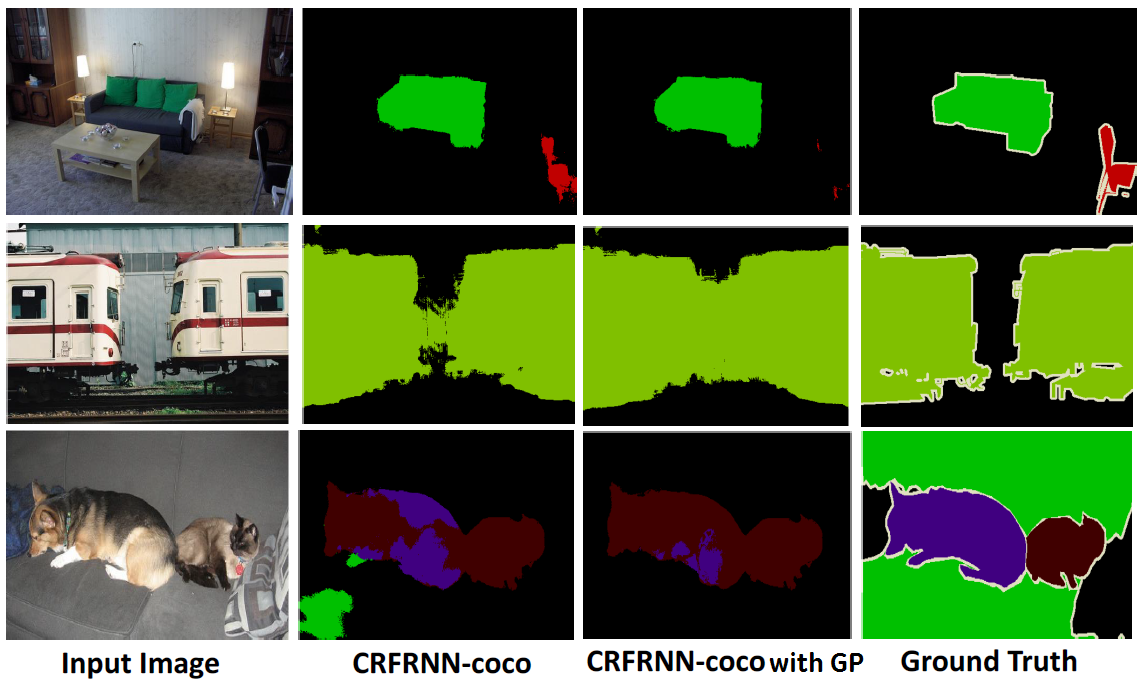}\label{d}}
\newline
\caption{Qualitative results on the PASCAL VOC2012 reduced validation set - Comparison with CRFRNN-coco pretrained model. Top half shows the successful outputs, Bottom half shows the failure cases.}
 \label{fig:crfrnn_coco_outputs_app}
\end{figure}

\begin{figure}
 \includegraphics[width=\textwidth]{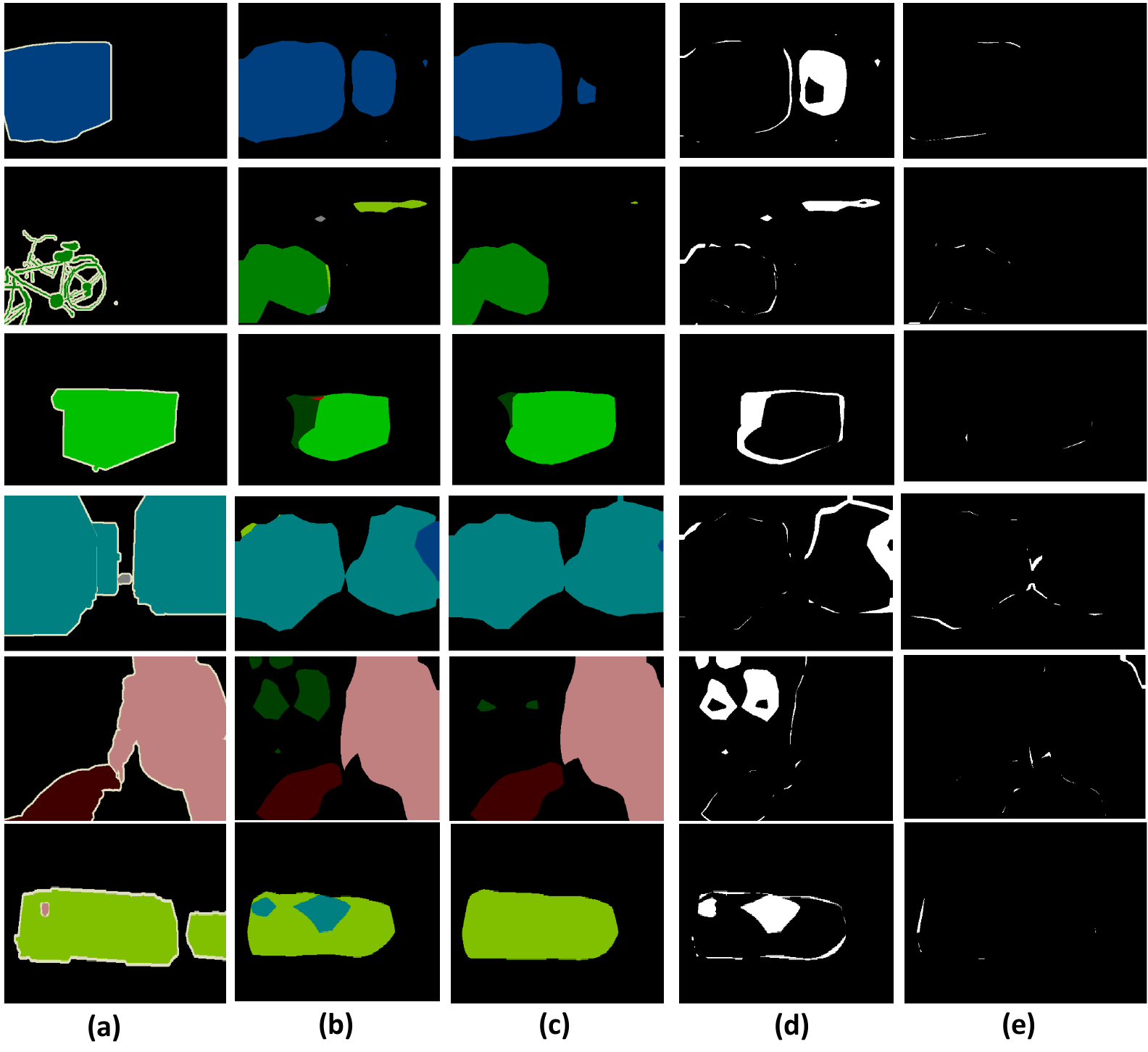}
 \caption{(a) Ground truth (b) Output of FCN-32s network (c) Output from the proposed approach (d) Pixels that were
incorrectly classified by FCN-32s corrected by our approach (e) Pixels that were incorrectly classified by our approach that
FCN-32s classified correctly.}
 \label{fig:boundary_examples}
 \end{figure}

\begin{figure}
 \includegraphics[width=\textwidth]{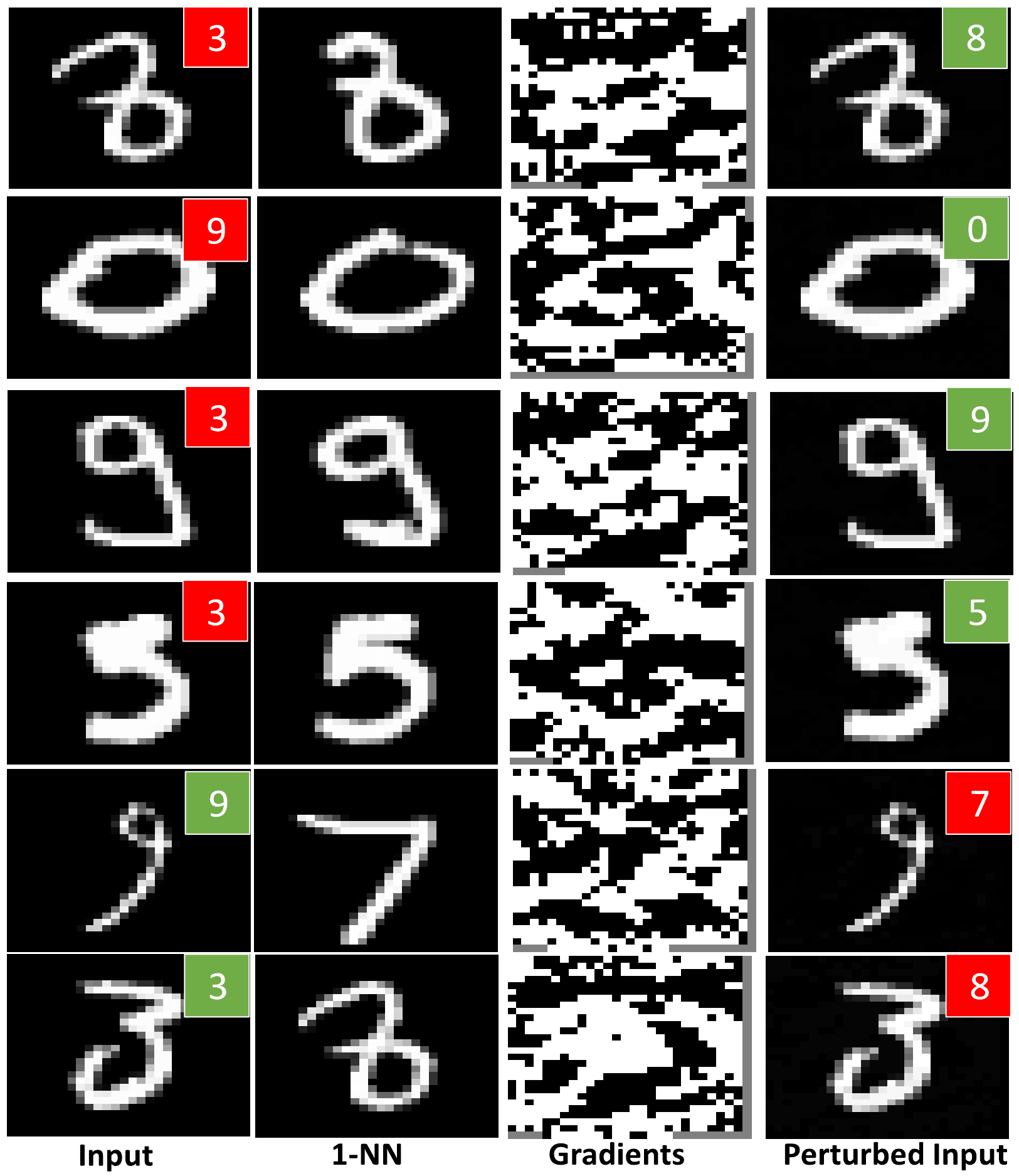}
 \caption{Example results of using the proposed approach for MNIST digits classification task. Top four rows shows situations
where our approach was successful in correcting the classifier errors while bottom two rows showcase the failures. The
red and green labels show the final deep network output: red indicates a mistake and green indicates correct prediction.}
 \label{fig:cls_app}
 \end{figure}

Appendix contains additional material and examples to support our paper. The explicit formula used in the error gradient computation for the toy example in Figure 10 from the paper is derived in section~\ref{sec:grad}. Figures~\ref{fig:fcn8s_outputs_app}, ~\ref{fig:fcn8s_coco_outputs_app} and ~\ref{fig:crfrnn_coco_outputs_app} show additional examples of improved performance when the proposed approach is used with the FCN-8s~\cite{fcncvpr}, FCN8s-coco~\cite{crfasrnn} and CRFasRNN~\cite{crfasrnn} pretrained models respectively. Figure~\ref{fig:boundary_examples} shows examples to supplement the claim made in Figure 4 in the paper that the pixels that are predicted correctly by our approach are more internal to the image whereas the small number of pixels that are predicted wrongly tend to occur towards the boundaries. These examples are generated using the FCN-32s deep network [14]. Finally, Figure~\ref{fig:cls_app} shows additional results of using our approach for the MNIST classification task.


\vspace{4mm}
\section{Error gradient computation for Figure~\ref{fig:unifGP}}\label{sec:grad}
Let the score output of the deep network be: $\mathbf{z} \in \mathbb{R}^{N_c}$. To get a probability distribution over classes, this is passed through a softmax operator whose output is given as: $\mathbf{y}=\big\{\frac{e^{z_i}}{\sum_{i}e^{z_i}}\big\}_{i=1}^{N_c}$, where $N_c$ is the number of classes. If $k \in [1,N_c]$ is the correct class, then the error gradient computed at the softmax output with respect to its input $\mathbf{z}$ is given as follows: Let $\sum_C$ denote $\{\sum_{i=1}^{N_c}e^{z_i}\}$, then
\begin{align}
\text{ if $i$ = $k$}: & \quad \frac{\partial y_i}{\partial z_i}=\frac{\sum_C.{e^{z_i}}-e^{z_i}.e^{z_i}}{\sum_C^2}=\frac{e^{z_i}}{\sum_C} \Bigg(1-\frac{e^{z_i}}{\sum_C}\Bigg)=y_i(1-y_i)=y_k(1-y_i) \\
\text{ if $i$ $\neq$ $k$}: & \quad  \frac{\partial y_i}{\partial z_k}=-\frac{0-e^{z_i}.e^{z_k}}{\sum_C^2}=-\frac{e^{z_i}}{\sum_C} \frac{e^{z_k}}{\sum_C}=-y_i y_k= y_k(0-y_i)
\end{align}

(2)-(3) could be summarized in the following single equation: 
\begin{align}
\frac{\partial \mathbf{y}}{\partial \mathbf{z}}= y_k(\Bell-\mathbf{y})
\end{align}

where $\Bell \in  \mathbb{R}^{N_c}$ is the label distribution, which in this case is a one hot vector with $l_k = 1$ and others zero. For a more general case, where $\Bell$ defines a distribution among classes, this formula generalizes in a straight forward manner as follows:

\begin{align}
\frac{\partial \mathbf{y}}{\partial \mathbf{z}}= (\Bell\cdot\mathbf{y})(\Bell-\mathbf{y})
\end{align}

It can be observed that (5) is a general version of (4) since the maximum probability value $y_k$ is replaced by the dot product between the label distribution and the output of softmax operation. 


\newpage 

\newpage 

\newpage 

\newpage

\newpage

\end{document}